\documentclass[10pt,twocolumn,letterpaper]{article}

\pdfoutput=1

\usepackage{graphicx}
\usepackage[margin=0.5in]{geometry}
\usepackage{amsmath}
\usepackage{amssymb}
\usepackage{booktabs}
\usepackage{multicol}
\usepackage{multirow}
\usepackage{sidecap}
\usepackage{subcaption}
\usepackage{tikz}

\usepackage[pagebackref,breaklinks,colorlinks]{hyperref}

\usepackage[capitalize]{cleveref}
\crefname{section}{Sec.}{Secs.}
\Crefname{section}{Section}{Sections}
\Crefname{table}{Table}{Tables}
\crefname{table}{Tab.}{Tabs.}

\renewcommand{\paragraph}[1]{\vspace{0.4em}\noindent\textbf{#1}}

\definecolor{vtab_natural}{rgb}{0.7137,0.3333,0.3333}
\definecolor{vtab_specialized}{rgb}{0.4118,0.6431,0.4314}
\definecolor{vtab_structured}{rgb}{0.3254,0.4431,0.6666}

\newcommand*\rot{\rotatebox{90}}
\newcommand{\eat}[1]{}

\begin{document}

\date{}
\title{Masked Autoencoding Does Not Help Natural Language Supervision at Scale}

\author{Floris Weers \and Vaishaal Shankar \and Angelos Katharopoulos \and Yinfei Yang \and Tom Gunter \and \\
\vspace{0.3cm}
{\tt\small \{fweers,vaishaal\_shankar,a\_katharopoulos,yinfei\_yang,tom\_gunter\}@apple.com}} 
\maketitle

\begin{abstract}
   Self supervision and natural language supervision have emerged as two exciting ways to train general purpose image encoders which excel at a variety of downstream tasks. Recent works such as M3AE \cite{geng_m3ae_2022} and SLIP \cite{mu_slip_2021} have suggested that these approaches can be effectively combined, but most notably their results use small ($<$20M examples) pre-training datasets and don’t effectively reflect the large-scale regime ($>$100M samples) that is commonly used for these approaches. Here we investigate whether a similar approach can be effective when trained with a much larger amount of data. We find that a combination of two state of the art approaches: masked auto-encoders, MAE \cite{he_mae_2021} and contrastive language image pre-training, CLIP \cite{radford_clip_2021} provides a benefit over CLIP when trained on a corpus of 11.3M image-text pairs, but little to no benefit (as evaluated on a suite of common vision tasks) over CLIP when trained on a large corpus of 1.4B images. Our work provides some much needed clarity into the effectiveness (or lack thereof) of self supervision for large-scale image-text training. 
\end{abstract}

\section{Introduction}
\label{sec:intro}

Large scale pretraining has become a powerful tool in the arsenal of computer vision researchers to produce state of the art results across a wider variety of tasks \cite{coca, beit, Hu_2022_CVPR, focalnets}. However, when pre-training on tens of millions to billions of images it is difficult to rely on standard supervised methods to train models, as datasets of this size often lack reliable labels.
In the presence of these massive but largely under-curated datasets, two general classes of methods to train general purpose image encoders have emerged:
\begin{enumerate}
    \item  \textit{Self Supervised} techniques that learn visual representations from the image data alone \cite{maes, simclr}
    \item  \textit{Natural Language Supervised} methods that utilize  paired free-form text data to learn visual representations \cite{clip, align}
\end{enumerate}

Due to the unique strengths and weaknesses of each approach\footnote{Self supervised methods can learn representations without labels, but natural language supervision learns better representations. Natural language supervised methods rely on quality of captions}, a recent flurry of work has introduced methods that combine both forms of supervision \cite{slip, declip, m3ae, flava} to varying degrees of success.  While each of these methods establishes some regime where the additional supervision helps, none of these ``joint-supervision'' methods advance state of the art in any meaningful way. Additionally, to our knowledge none of these methods have shown comparative results at the scale many large scale vision models are currently trained at ($>$100M examples) \cite{clip, align, Sun_2017_ICCV, basic, instagram2, laion5b, coca}.
Furthermore, methods that use both forms of supervision start with the presumption that the additional supervision \textbf{is helpful} and either often lack clean ablations or lack evaluations in a ``high accuracy'' regime---leading to further confusion regarding whether a combination of these methods can actually improve the state of the art.  To clarify this issue, in this work, we investigate a simple question: 

\begin{quote}
\textit{Does a combination of self supervision and natural language supervision actually lead to higher quality visual representations?}
\end{quote}

In order to answer this, we first introduce a straightforward baseline approach that combines standard self supervision and language supervision techniques. We combine masked auto-encoders (MAE) and contrastive language image-pretraining (CLIP) to make MAE-CLIP. We then present a careful study of the performance of MAE, M3AE, CLIP and MAE-CLIP across a wide variety of tasks in two distinct regimes: a ``low-sample" \footnote{We note that what low sample means has changed substantially over the last few years} 11.3 million example regime and a ``high-sample" 1.4 billion example regime. We train self-supervised and language-supervised methods using the same pre-training datasets under the assumption that we have no knowledge about downstream tasks. Our experiments show:
\begin{enumerate}
    \item In the low sample size regime, without changing the final pooling operation in the network, we observe a large performance improvement, namely $6$\% on ImageNet \cite{imagenet} and 4\% on VTAB \cite{vtab}. However, when we modify the pooling operation, the improvement \textbf{substantially} decreases to around $1$\% on both ImageNet and VTAB. 
    \item In the high sample size regime, there is virtually no difference in performance between MAE-CLIP and CLIP across ImageNet, VTAB, and VQA tasks. 
\end{enumerate}

We believe our work is the first careful study of this form and contextualizes recent progress in both self-supervision and natural language supervision.

The rest of the paper is organized as follows: In \Cref{sec:related_work}, we cover related work in the areas of self supervision and natural language supervision. In \Cref{sec:background}, we give an overview of the baseline methods we study, MAE, M3AE, CLIP and our new baseline MAE-CLIP. Then we present and analyse our small scale and large scale experimental findings in Sections \ref{sec:exp} and \ref{sec:exp_scale}. Finally, we discuss potential explanations for our findings and some future work in \ref{sec:analysis_discussion}.

\section{Related Work}
\label{sec:related_work}

\begin{figure*}
    \begin{center}
	\includegraphics[width=0.8\textwidth]{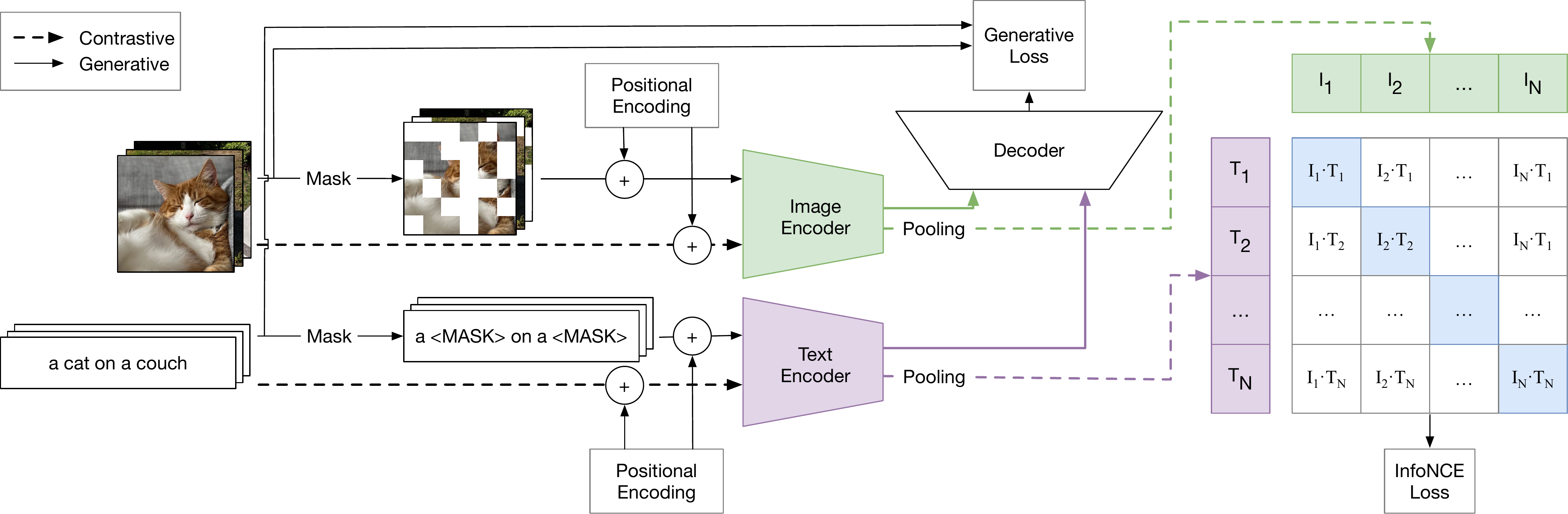}
    \end{center}
    \caption{\textbf{MAE-CLIP}: CLIP \cite{radford_clip_2021} augmented with a generative decoder in the style of MAE \cite{geng_m3ae_2022}.}
    \label{fig:architecture}
    \vspace{-0.5cm}
\end{figure*}

Our work combines \textbf{natural language supervision} and \textbf{self-supervision} in a \textbf{multi-task} approach to visual encoding, and so research from these three areas is relevant.

\textbf{Natural language supervision} for visual encoding covers a variety of approaches that assume access to datasets of images or videos associated with text.
Some of the most successful use a contrastive pairwise alignment signal applied to very large batch and dataset sizes \cite{radford_clip_2021}. This large batch size means hard negative pairs can be produced by random sampling, avoiding the need for a memory bank \cite{chen_simclrv2_2020, chen_mocov2_2020, he_mocov1_2020} or momentum distillation \cite{chen_mocov3_2021}.
FILIP \cite{yao_filip_2021} further improves zero-shot performance for some tasks by using a more fine-grained elementwise contrastive loss.
Image captioning has also shown promise as a pre-training task \cite{desai_virtex_2021, wang_simvlm_2021}, with \cite{desai_virtex_2021} in particular demonstrating strong data efficiency. UNITER \cite{chen_uniter_2020} pursues a similar idea, applying a generative loss to both image and text modalities.
\cite{tan_lxmert_2019, dou_meter_2021, arici_mlim_2021,yu_ernie-vil_2020,li_semvlp_2021, lu_vilbert_2019,su_vl-bert_2020,chen_uniter_2020,huang_seeing_2021, kim_vilt_2021} apply the masked patch prediction problem from \cite{devlin_bert_2019, lewis_bart_2019, he_mae_2021} to a joint image-text data space. These approaches either use a pre-trained convolutional neural network (CNN) to generate region of interest (RoI) proposal encodings, or predict the labelled class of the masked patch instead of the raw pixel values (or the quantized patch ID).
UNIMO-2 \cite{li_unimo-2_2022} attempts to ground image and text patches for a single data example into a top-$k$ quantized set of embeddings, before passing the grounded embeddings along with the raw non-grounded embeddings into a decoder, targeting a loss based on masked language modeling (MLM) and image-text-matching (ITM).

\textbf{Self-supervised} approaches for visual encoding may be loosely categorized into those that target consistency constraints between multiple views of the same scene and those that attempt visual reconstruction on corrupted or conditioning representations of images.
Consistency constraints are often derived through data-augmentation applied to a single real image \cite{zbontar_barlow_2021, chen_simclr_2020, grill_byol_2020, chen_simsiam_2020, caron_dino_2021}, but these may also take the form of e.g. a temporal constraint if video data is available \cite{wang_unsupervised_2015}, or be applied at an image-patch level by making a smoothness assumption and selecting neighboring patches as `consistent' whilst treating distant patches (or those sourced from another image) as negative examples \cite{yun_patch-level_2022}.
Forms of denoising autoencoder (DAE) \cite{vincent_extracting_2008} are popular as a means of self-supervision, and have been investigated at a variety of scales; iGPT \cite{chen_igpt_2020} learns to generate images as a row-major flattened sequence of pixels, whilst \cite{oord_representation_2019}, \cite{razavi_generating_2019}, and \cite{esser_vqgan_2021} compress images into a short sequence of discrete codes before then regenerating and scoring at a pixel level. BEiT \cite{bao_beit_2021} first encodes an image into a sequence of discrete codes before masking and then predicting a subset of the codes. MAE \cite{he_mae_2021} drops the discretization, instead predicting raw pixel loss for a subset of the encoded patches in a manner strongly reminiscent of BART \cite{lewis_bart_2019}. SplitMask \cite{el-nouby_splitmask_2021} applies both a masked image patch prediction loss and a pairwise contrastive loss, using the masked example as well as the inverted masked example to form each positive pair. Meanwhile, M3AE \cite{duan_multi-modal_2022} and VL-BEiT \cite{bao_vl-beit_2022} both propose a masked patch prediction problem applied to image and text modalities jointly. Recent concurrent work shows that while masking does \emph{not} help generalization, the masking can be used to speed up training by dropping the masked tokens during the forward and backward pass \cite{scalemask}.  EVA \cite{eva} shows that doing masked image modeling to predict the output \emph{embeddings} (rather than pixels) of the masked image can be an effective pretraining task when combined with natural language supervision. We note for EVA the masked image modeling occurs only on smaller well curated datasets such as ImageNet, Conceptual Captions and COCO.

\textbf{Multi-task} methods for pre-training visual encoders are a highly active area of research.
CoCA \cite{yu_coca_2022} combines the cross-modality contrastive task from \cite{radford_clip_2021} with image captioning in the style of \cite{desai_virtex_2021}, and by pre-training at very large scale show that the resulting model is more performant than prior art across a very broad array of downstream visual understanding tasks.
SLIP \cite{slip} combines a SimCLR style self supervision loss with the standard CLIP contrastive loss to train a jointy self-supervised and natural language supervised model, however while they show large performance gains they are all in a low accuracy regime (below 40\% Top-1 accuracy on ImageNet). 
Florence \cite{yuan_florence_2021} employs a diverse array of tasks and datasets at pre-training time (including object detection, CLIP-style contrastive modeling, Visual Question Answering (VQA), and supervised classification).
ALBEF \cite{li_align_2021} builds on a CLIP-style architecture by incorporating a single-stream encoder that consumes both modalities. It uses a masked language modeling loss for the text modality and an image-text matching loss for within-batch hard-negatives, mined according to contrastive similarity.
FLAVA \cite{singh_flava_2022} also uses masked-image modeling, and uses both single and joint modality encoders, with additional single-modality datasets (i.e. not just image-text pairs) that allow it to generalize to longer text inputs and visual data which is unlikely to be captioned.
X-VLM \cite{zeng_xvlm_2022} utilizes image data which includes object-labelled bounding boxes to extend the image-text matching and contrastive losses to cover within-image regions as well as whole-image captions. They also include a masked language modelling objective, achieving impressive performance on many zero-shot downstream tasks---although the requirement for object bounding box data may not scale well for large pre-training datasets. 
SupMAE \cite{liang_supmae_2022} adds a supervised labeling task to the MAE architecture, demonstrating that this results in better data efficiency.
MVP \cite{wei_mvp_2022}, meanwhile, finetunes a pre-trained CLIP backbone using MIM, and demonstrates that starting with a pre-trained model improves downstream visual recognition performance.
\textbf{MAE-CLIP} is reminiscent of M3AE (with the addition of a contrastive task), ALBEF (but applying the masked prediction task to both modalities, dropping the momentum distillation), and most of all FLAVA (without the single-modality masked-patch-prediction tasks, and with the use of high ratio random masking following MAE and M3AE).
\vspace{-0.3cm}
\section{Background}
\label{sec:background}

\subsection{Contrastive language-image pre-training}

CLIP \cite{radford_clip_2021} and ALIGN \cite{jia_align_2021} both demonstrated that contrastive image-language supervision applied at scale was capable of producing an image encoder that excels at a range of downstream tasks---often in a zero-shot setting.
In our implementation, we begin by following the design outlined in \cite{radford_clip_2021}.
Specifically, we use per-modality encoders to produce a single $\ell2$-normalized dense embedding vector for each image or text input, before applying a pairwise InfoNCE \cite{oord_representation_2019} loss to a large global batch of paired image-text embeddings, using all non-paired examples as hard negatives.
We define the image-to-text loss $L_{i2t}$ and the text-to-image loss $L_{t2i}$ as:
\begin{equation}
    L_{i2t} = -\frac{1}{N}\sum_{j}^{N}\log {\frac{\exp (x_j^T y_j/\sigma)}{\sum_{k=1}^{N}\exp(x_j^T y_k/\sigma)}} \label{eqn:clip_loss}
\end{equation}
\begin{equation}
    L_{t2i} = -\frac{1}{N}\sum_{j}^{N}\log {\frac{\exp (y_j^T x_j/\sigma)}{\sum_{k=1}^{N}\exp(y_j^T x_k/\sigma)}}, \label{eqn:clip_loss_t2i}
\end{equation}
where $N$ is the global batch size, $x_j$ is the normalized embedding of the image for the $j$-th pair and $y_k$ is the normalized embedding of the text in the $k$-th pair in the batch. $\sigma$ is a learnable temperature parameter. The total contrastive loss $L_c = \frac{1}{2}(L_{t2i} + L_{i2t})$ is the average of these two losses.

We use transformer encoders for both modalities, however, differently to \cite{radford_clip_2021} we explore several strategies for aggregating the image-patch or text-token outputs into a single embedding (see Section \ref{sec:pooling_analysis}). We also eschew the autoregressive masking strategy for the text encoder, instead allowing full bi-directional self attention following \cite{jia_align_2021}.

\subsubsection{Pooling}
In CLIP \cite{radford_clip_2021} the authors use a separate `CLS' token projected through the network as the overall image representation.  On the other hand, ALIGN \cite{jia_align_2021} use global-average pooling over the encoded visual features.  Whilst both approaches produced good downstream evaluation results, recent work \cite{chen2021vseinfty,zhai_2022_cvpr,ranasinghe2023perceptual}, suggested that the choice of pooling strategy can strongly influence the quality of visual semantic embeddings. In particular, \cite{ranasinghe2023perceptual} demonstrated that this effect is present for CLIP-like visual encoders. Noting this, we opt to investigate three pooling strategies: 1. the default multihead-attention pooling (MAP), 2. global average pooling (GAP)---both described in \cite{zhai_2022_cvpr}---as well as 3. non-maximal suppression pooling (MAX) as in \cite{chen2021vseinfty,ranasinghe2023perceptual}.

\subsection{Masked Autoencoders}
In MAE \cite{he_mae_2021} the authors demonstrate a simple technique for self-supervised image-encoder pre-training that---to our knowledge---is still  considered state-of-the-art. They use a ViT \cite{dosovitskiy_vit_2021} encoder-decoder architecture and apply it to heavily masked input images.
The input to the encoder consists of the visible, unmasked image patches, first embedded via a linear projection before additive positional embeddings are applied, and then the result is fed through the encoder's transformer layers. They demonstrate that a very high patch masking ratio is critical to achieving good performance, and usually retain only $25\%$ of the image.
The decoder consumes the output of the encoder, as well as a learned `masked-patch' embedding, which is included to represent each masked token. Positional embeddings are also reapplied, to ensure that spatial information is passed through the decoder for the masked patches. The output of the decoder at the masked positions is then measured against ground truth using a simple mean-squared-error loss. As in \cite{he_mae_2021} we experiment with predicting both normalized and un-normalized patch values, finding that predicting the normalized patch value slightly improves the performance of our MAE implementation.

\textbf{Masking across modalities:}
BART \cite{lewis_bart_2019} applies a similar strategy to text data, but uses a cross-entropy loss over the masked-token output distribution and a far lower masking ratio (typically around $15\%$).  M3AE \cite{geng_m3ae_2022} extends MAE and BART to incorporate inputs from both text and image modalities, using per-modality input and output projections. Otherwise the encoder-decoder architecture resembles MAE, with the addition of learned modality-indicating embeddings to each transformer input.

The M3AE loss relies only on the contents of the image-text pairs, and comprises mean-squared error for the masked image patches ($L_{gen\_i}$) and cross-entropy over the vocabulary for the masked text tokens ($L_{gen\_t}$):

\begin{align}
    L_{\text{gen\_i}} & =  \frac{1}{N}\sum_{j}^{N} \left( p_j - P_j \right)^2 \label{eqn:m3ae_loss} \\
    L_{\text{gen\_t}} & = - \frac{1}{N}\sum_{j}^{N} \log \frac{\exp(t_{n,T_n})}{\sum_{c=1}^C \exp(t_{n,c})}, \label{eqn:m3ae_loss_gen_t}
\end{align}

where $p_j$ and $P_j$ refer to the predicted and ground truth pixel value respectively and $t_n$ and $T_n$ refer to the predicted token distribution and the ground truth token, with $C$ as the number of unique tokens. For the image reconstruction loss, we normalize the ground-truth per-patch, following \cite{he_mae_2021}.

\subsection{MAE-CLIP}
As the name suggests, MAE-CLIP attempts to incorporate aspects of MAE/M3AE into CLIP through the addition of a single dual-modality transformer decoder (as in M3AE) which consumes the output of the CLIP encoders. We show this in \Cref{fig:architecture}.

\subsubsection{Components}

The model architecture is similar to ALBEF~\cite{li_align_2021} and FLAVA~\cite{flava}, consisting of three components. 

\textbf{An image encoder:}
Following ViT \cite{dosovitskiy_vit_2021}, we divide the input image into equally-sized, non-overlapping patches. After applying a linear projection and adding a 2-D position encoding, we feed the per-patch representations through a number of transformer layers \cite{vaswani_attention_2017}.

\textbf{A text encoder:}
Following ALIGN \cite{align}, our text encoder is based on BERT \cite{devlin_bert_2019}, in which text is first tokenized and embedded, and a $1$-D trainable position encoding is added. Differently to BERT, we use pre-layernorm \cite{xiong_pre-layernorm_2020} and initialize the parameters according to the simplified scheme outlined in \cite{radford_gpt-2_2019}.

\textbf{A cross-modality decoder:}
The decoder receives per-element encoded image and text representations from the encoders for both masked and un-masked elements. For the image modality, masked patches are added by replacing their values with a shared trainable mask token. Positional encodings are once again added to all elements, following \cite{he_mae_2021}. We also add a per-modality trainable encoding to allow the decoder to easily distinguish between the two modalities. The decoder uses the same transformer implementation as the encoders. The output of the final decoder layer is then projected into a per-patch-modality output space and the overall loss is computed.

\vspace{-0.1cm}
\subsubsection{Losses}
\label{sec:mae_clip_loss}
%
%

MAE-CLIP employs the losses of both CLIP (Eq. \ref{eqn:clip_loss}, \ref{eqn:clip_loss_t2i}) and M3AE (Eq. \ref{eqn:m3ae_loss}, \ref{eqn:m3ae_loss_gen_t}). Our final loss is a weighted sum of the losses from each task, as follows:
\begin{equation}
\label{qe:mae_clip_loss}
L = \frac{1}{2}\left(L_{t2i} + L_{i2t}\right) + w_{i} \cdot L_{\text{gen\_i}} + w_{t} \cdot L_{\text{gen\_t}}
\end{equation}
where $w_{i}$ and $w_{t}$ are scalars used to control the relative weight of the generative losses. We always provide paired image-text inputs to the model, leaving it to future work to explore the benefits of also incorporating single modality inputs and reconstruction losses, at scale.

To avoid severely impacting the CLIP loss via computing it with masked inputs, we run the encoders twice: once with full unmasked input in order to compute the CLIP loss, before then making a second pass using only the unmasked input to compute the M3AE loss. Finally, we compute the weighted sum of the two losses, and use this to calculate the overall update. \footnote{Alternative, more memory efficient, strategies exist for computing the overall updates (e.g. calculate per-loss gradients, or perform round-robin updates), but prior work \cite{alayrac_flamingo_2022} has demonstrated that some of these techniques may negatively impact the overall training result.}

\section{Experiments}
\label{sec:exp}

In this section, we present our experimental results to study how the addition of self-supervision affects the representation quality of natural language supervised models.  We study standard masked auto-encoders (MAE), multimoddal masked auto-encoders (M3AE), contrastive language image pretraining (CLIP) and our newly introduced baseline method (MAE-CLIP), which combines mask based self-supervision with contrastive language-image learning.


\subsection{Experimental Setup}
\label{sec:experimental_setup}
\textbf{Architecture:}
Experiments are performed using an image encoder based on the architecture described as ViT-B-16 in \cite{xiao_early_2021}, in combination with a $12$-layer, $512$-wide text encoder with $8$ attention heads. For the decoder, we use another $12$-layer, $512$-wide transformer  with $8$ attention heads, identical to the text encoder.
We also use a byte-pair encoding tokenizer, fit to the OpenWebText corpus~\cite{openwebtext_corpus}. When training MAE-CLIP, we replace the random masking by a similarity masking strategy, which makes use of the element-wise CLIP similarity scores to select the masked element for the reconstruction task.
\footnote{We compare the similarity masking strategy with the original random masking strategy on various tasks, the results show they perform similarly.}

\textbf{Pretraining Datasets:}
Our analysis is divided into two sections. First, in sections \ref{sec:image_classification} to \ref{sec:pooling_analysis}, we present our study on the ``low-sample'' regime using the combination of CC12M \cite{changpinyo_conceptual_2021} and CC3M \cite{sharma_gcc_2018} excluding all images whose text contains a \textit{[PERSON]} tag\footnote{Preliminary experiments on CLIP suggest removing such data leads to better performance.}. The final dataset contains ~11.3 million comparatively high quality image-text pairs.
We refer to this dataset as \emph{CC} or ``small'' scale.

\textbf{Pretraining configuration:}
We use a minibatch size of $16,384$ samples and train using the AdamW optimizer \cite{loshchilov_adamw_2019} with decoupled weight decay. We use a base learning rate of $5\times10^{-4}$ and train using cosine learning rate decay \cite{loshchilov_sgdr_2017} with an initial linear warmup of $200$ steps. We reduce the $\beta_2$ decay factor to $0.98$, and increase $\epsilon$ to $10^{-6}$, finding that these changes improved rate of convergence for all models trained. To increase training convergence speed, we switch from a local (per-GPU) to a global contrastive loss after $500$ steps. During the global contrastive loss phase, we set the generative text and image loss weights to $w_t=0.05$ and $w_i=0.1$ respectively. For all models which rely on masking, we use a $75\%$ masking ratio, consistent with \cite{he_mae_2021,geng_m3ae_2022}, as we did not find an alternative that improved downstream results. In total, we train our models on \emph{CC} for $32$ epochs, which corresponds to $\sim 22,000$ steps.
For our M3AE training runs, we use per-modality encoders so as to produce a baseline that is flop and parameter matched to our MAE-CLIP runs.


\begin{table}[t]
\small
    \centering
    \begin{tabular}{lcc}
    \toprule
        \textbf{Models} & \textbf{Zero-shot} & \textbf{Linear Probing} \\ 
        \midrule
        MAE          &   -- & 33.9 \\
        M3AE         &   -- & 52.5\\
        \midrule
        CLIP & 29.7 & 52.6 \\
        MAE-CLIP   & 33.8 & 58.9 \\ 
        \bottomrule
    \end{tabular}
    \caption{ImageNet classification with zero-shot transfer or linear probing after pretraining on the \emph{CC} dataset (11.3M images). MAE-CLIP significantly improves the classification performance of CLIP in the small scale regime.}
    \label{tab:cc_imagenet}
    \vspace{-0.5cm}
\end{table}

\begin{table*}[t]
\caption{Linear probing accuracy (\%) on classification tasks. All models are trained on the \emph{CC} Dataset (11.3M images). ({\color{vtab_natural}$\bullet$} VTAB/natural, {\color{vtab_specialized}$\bullet$} VTAB/specialized and {\color{vtab_structured}$\bullet$} VTAB/structured.)}
\label{tab:cc_linear_probe}
\small
\centering
\setlength{\tabcolsep}{3.2pt}
\begin{tabular}{@{}l|ccccccccccccccccc|c@{}}
&
  \rot{{\color{vtab_natural}$\bullet$} Caltech101} &
  \rot{{\color{vtab_natural}$\bullet$} CIFAR-100} &
  \rot{{\color{vtab_natural}$\bullet$} DTD} &
  \rot{{\color{vtab_natural}$\bullet$} Flowers102} &
  \rot{{\color{vtab_natural}$\bullet$} Pets} &
  \multicolumn{1}{c|}{\rot{{\color{vtab_natural}$\bullet$} SVHN}} &
  \rot{{\color{vtab_specialized}$\bullet$} EuroSAT} &
  \rot{{\color{vtab_specialized}$\bullet$} Camelyon} &
  \multicolumn{1}{c|}{\rot{{\color{vtab_specialized}$\bullet$} Resisc45}} &
  \rot{{\color{vtab_structured}$\bullet$} Clevr/Closest} &
  \rot{{\color{vtab_structured}$\bullet$} Clevr/Count} &
  \rot{{\color{vtab_structured}$\bullet$} DMLab} &
  \rot{{\color{vtab_structured}$\bullet$} dSprites/Ori} &
  \rot{{\color{vtab_structured}$\bullet$} dSprites/Loc} &
  \rot{{\color{vtab_structured}$\bullet$} KITTI/Dist} &
  \rot{{\color{vtab_structured}$\bullet$} sNORB/Azim} &
  \multicolumn{1}{c|}{\rot{{\color{vtab_structured}$\bullet$} sNORB/Elev}} &
  \rot{Average} \\
  \midrule
    MAE
        & 75.3 & 56.2 & 58.5 & 70.2 & 37.4 & 71.9 & 96.2 & 82.8 & 84.0
        & 65.1 & 60.6 & 42.8 & 37.9 & 79.9 & 36.7 & 37.4 & 64.5 & 62.2 \\
    M3AE
        & 85.6 & 66.6 & 69.4 & 86.5 & 58.4 & 69.2 & 97.4 & 82.9 & 90.7
        & 69.4 & 71.2 & 48.2 & 49.8 & 83.3 & 42.6 & 34.3 & 73.4 & 69.3  \\
    CLIP
        & 84.2 & 62.8 & 57.7 & 81.6 & 69.7 & 52.5 & 95.5 & 82.6 & 86.7
        & 53.7 & 52.9 & 44.8 & 45.9 & 61.3 & 45.7 & 31.0 & 42.6 & 61.8  \\
    MAE-CLIP
        & 89.8 & 66.2 & 64.8 & 86.5 & 74.5 & 56.7 & 95.4 & 81.6 & 86.9
        & 53.6 & 59.6 & 45.9 & 44.6 & 68.6 & 48.0 & 36.1 & 51.0 & 65.3 \\
    \bottomrule
\end{tabular}
\end{table*}

\begin{table}[t]
\small
    \centering
    \begin{tabular}{lccc}
        \toprule
        \textbf{Model} & \textbf{CLEVR} & \textbf{VQAv2} & \textbf{GQA} \\
        \midrule
        MAE & 93.5 & 45.7 & 46.1 \\
        M3AE & \textbf{97.5} & 55.8 & 50.9 \\
        CLIP & 87.5 & 55.6 & 50.0 \\
        MAE-CLIP & 96.0 & \textbf{58.5} & 52.2 \\
        \bottomrule
    \end{tabular}
    \caption{VQA finetuning results for models pre-trained on CC Dataset (11.3M images). We train a new decoder for all methods while keeping the encoders frozen. MAE-CLIP performs significantly better than either only self-supervised or language supervised methods by themselves.}
    \label{tab:cc_vqa}
\end{table}

\subsection{Image Classification} \label{sec:image_classification}

We evaluate the quality of the representations learned from pre-training on \emph{CC} by measuring the image classification performance, either by zero-shot transfer, or by training a linear classifier using the predicted visual features. For zero-shot classification, we use the average of $80$ prompts and follow the routine prescribed in \cite{clip}. To train the linear classifier, we pre-compute features using the visual encoder\footnote{We use no data augmentation here} and run AdamW for $20$ to $80$ epochs with a learning rate of $0.01$.

\Cref{tab:cc_imagenet} shows the zero-shot and linear probing classification results on the ImageNet \cite{deng_imagenet_2009} dataset while \Cref{tab:cc_linear_probe} shows our linear probing results on most ($17$) of the VTAB \cite{zhai_vtab_2020} benchmark datasets\footnote{We do not include results for either the diabetic retinopathy or the Sun397 tasks, due to licensing issues. More details in the supplementary material}. Initially, we observe that the combination of self-supervision and natural language supervision provides a consistent and substantial improvement over either form of supervision by itself. This result concurs with previous works such as \cite{m3ae} and \cite{slip}, which show that self-supervision aids natural language supervision.

\subsection{VQA} \label{sec:cc_vqa}

Subsequently, we compare the performance of MAE-CLIP to the rest of the baselines on the visual question answering task using three datasets. CLEVR \cite{johnson_clevr_2017}, VQAv2 \cite{goyal_vqa2_2017} and GQA \cite{hudson_gqa_2019}. VQA assesses the model's multimodal reasoning capabilities as well as its visual grounding. In order to finetune our models for VQA, we freeze the image and text encoders and either randomly reinitialize the decoder in models such as MAE-CLIP or M3AE or add a new identical decoder for CLIP. During finetuning, we use a layer-wise learning rate decay, following \cite{clark_electra_2020, he_mae_2021, bao_beit_2021}. For CLEVR and VQAv2 we finetune the decoder for 50 epochs, while for GQA we finetune for 5 epochs on the ``train-all'' split and 2 epochs on the ``train-balanced'' split following the protocol of \cite{li_oscar_2020}. In all cases, we treat the problem as a classification problem of selecting one answer out of the set of possible answers for each dataset.

In \Cref{tab:cc_vqa} we observe that self-supervision combined with natural language supervision (MAE-CLIP) performs consistently better than either MAE or CLIP by themselves. Combining information from both modalities is evidently critical for effectively answering visual questions as the performance of MAE on VQAv2 and GQA is several percentage points lower than the rest of the methods. Interestingly, methods that employ a form of self-supervision perform significantly better on CLEVR. We argue that this discrepancy is due to CLEVR requiring object localization and spatial reasoning which benefits from this form of self-supervision.

\subsection{Pooling Analysis} \label{sec:pooling_analysis}
 To better understand whether the gap between MAE-CLIP and CLIP we observed in the previous section is fundamental, we investigate the effect of the pooling operator on different downstream tasks. \Cref{tab:pooling} compares the performance of CLIP and MAE-CLIP with the standard multi-head attention pooling (MAP), global average pooling (GAP) and max pooling (MAX) on all image classification tasks and visual question answering. We note that GAP and MAX pooling perform substantially better than the standard MAP pooling across ImageNet, VTAB and VQA tasks. 
A hypothesis posed by \cite{ranasinghe2023perceptual} is that the alternative pooling operators lead to better \textit{perceptual grouping} in the visual representation. We study this with a qualitative analysis in Section \ref{sec:grounding}.

\begin{table}[t]
\small
    \centering
    \begin{tabular}{ll|ccccc}
    \toprule
        \multirow{2}*{\bf Model} & \multirow{2}*{\bf Pooling} & \multicolumn{2}{c}{\bf ImageNet} & \textbf{VTAB} & \textbf{VQA}\\
        &  & ZS & LP & LP & FT \\ 
        \midrule
        CLIP & MAP      & 29.7 & 52.6 & 61.8 & 64.4 \\
        MAE-CLIP & MAP  & 33.8 & 58.9 & 65.3 & 68.9 \\
        \midrule
        CLIP & GAP      & 29.3 & 59.8 & 70.6 & 65.1 \\
        MAE-CLIP & GAP  & 33.5 & 62.5 & \textbf{71.7} & \textbf{69.2} \\
        \midrule
        CLIP & MAX      & 33.1 & 62.3 & 68.7 & 63.5 \\
        MAE-CLIP & MAX  & \textbf{35.2} & \textbf{63.2} & 69.1 & 68.5 \\
        \bottomrule
    \end{tabular}
    \caption{Results of CLIP and MAE-CLIP with different pooling options trained on CC dataset (11.3M images). We compare the models on ImageNet Zero-Shot~(ZS), ImageNet Linear-Probing~(LP),  VTAB 17-task Linear-Probing average, and VQA 3 tasks Fine-Tuning~(FT) average.}
    \label{tab:pooling}
    
\end{table}

\begin{table*}[t]
\caption{Linear probing accuracy (\%) on classification tasks. Models are all trained on our \emph{web-crawled} dataset (1.4B images). ({\color{vtab_natural}$\bullet$} VTAB/natural, {\color{vtab_specialized}$\bullet$} VTAB/specialized and {\color{vtab_structured}$\bullet$} VTAB/structured.) In the large scale pretraining regime, the difference between MAE-CLIP and CLIP is reduced to $<1\%$. * At evaluation time, our M3AE model was $50\%$ trained, so that performance may improve further.}
\label{tab:large_linear_probe}
\small
\centering
\setlength{\tabcolsep}{3.2pt}
\begin{tabular}{@{}l|ccccccccccccccccc|c@{}}
&
  \rot{{\color{vtab_natural}$\bullet$} Caltech101} &
  \rot{{\color{vtab_natural}$\bullet$} CIFAR-100} &
  \rot{{\color{vtab_natural}$\bullet$} DTD} &
  \rot{{\color{vtab_natural}$\bullet$} Flowers102} &
  \rot{{\color{vtab_natural}$\bullet$} Pets} &
  \multicolumn{1}{c|}{\rot{{\color{vtab_natural}$\bullet$} SVHN}} &
  \rot{{\color{vtab_specialized}$\bullet$} EuroSAT} &
  \rot{{\color{vtab_specialized}$\bullet$} Camelyon} &
  \multicolumn{1}{c|}{\rot{{\color{vtab_specialized}$\bullet$} Resisc45}} &
  \rot{{\color{vtab_structured}$\bullet$} Clevr/Closest} &
  \rot{{\color{vtab_structured}$\bullet$} Clevr/Count} &
  \rot{{\color{vtab_structured}$\bullet$} DMLab} &
  \rot{{\color{vtab_structured}$\bullet$} dSprites/Ori} &
  \rot{{\color{vtab_structured}$\bullet$} dSprites/Loc} &
  \rot{{\color{vtab_structured}$\bullet$} KITTI/Dist} &
  \rot{{\color{vtab_structured}$\bullet$} sNORB/Azim} &
  \multicolumn{1}{c|}{\rot{{\color{vtab_structured}$\bullet$} sNORB/Elev}} &
  \rot{Average} \\
   \midrule
    M3AE*
        & 93.0 & 74.8 & 78.2 & 95.4 & 81.2 & 69.6 & 97.3 & 84.8 & 92.7
        & 65.7 & 74.7 & 51.0 & 52.7 & 80.7 & 47.5 & 36.2 & 68.8 & 73.2 \\
    CLIP
        & 94.9 & 78.4 & 80.0 & 97.3 & 86.9 & 59.0 & 94.1 & 82.3 & 92.7
        & 45.6 & 62.1 & 46.0 & 46.1 & 53.3 & 50.9 & 20.3 & 35.8 & 66.2  \\
    CLIP\textsubscript{MAX}
        & 96.1 & 81.0 & 80.9 & 97.3 & 89.9 & 65.7 & 96.0 & 83.2 & 94.1
        & 52.8 & 67.8 & 49.9 & 59.5 & 67.6 & 41.2 & 23.4 & 45.8 & 70.1 \\
    MAE-CLIP\textsubscript{MAX}
        & 95.8 & 79.2 & 81.5 & 96.8 & 88.2 & 62.1 & 95.8 & 81.8 & 93.0
        & 52.0 & 66.9 & 49.6 & 53.7 & 72.5 & 53.0 & 32.3 & 45.4 & 70.6 \\
    CLIP\textsubscript{GAP}
        & 95.8 & 80.5 & 81.6 & 97.6 & 88.7 & 66.0 & 97.0 & 84.4 & 93.3
        & 56.7 & 71.4 & 53.3 & 58.0 & 70.1 & 50.6 & 38.3 & 55.1 & 72.9  \\
    MAE-CLIP\textsubscript{GAP}
        & 95.4 & 79.3 & 82.2 & 97.4 & 88.6 & 72.8 & 96.6 & 84.5 & 93.5
        & 57.5 & 73.6 & 52.7 & 57.5 & 71.2 & 51.6 & 45.6 & 55.2 & 73.8 \\
    \bottomrule
\end{tabular}
\vspace{-0.2cm}
\end{table*}

\section{Experiments at Scale}
\label{sec:exp_scale}
\begin{table}
    \centering
    \begin{tabular}{lcc}
    \toprule
        \textbf{Models} & \textbf{Zero-shot} & \textbf{Linear Probing} \\ 
        \midrule
        M3AE$^*$       &   -- & 69.3 \\
        CLIP\textsubscript{GAP}       & 61.8 & 75.9 \\
        CLIP\textsubscript{MAX}       & 63.7 & 77.5 \\
        MAE-CLIP\textsubscript{GAP} & 57.4 & 75.7 \\ 
        MAE-CLIP\textsubscript{MAX} & 60.9 & 76.6 \\
        \bottomrule
    \end{tabular}
    \caption{ImageNet classification after pretraining on \emph{web-crawled} dataset (1.4B images). In the large scale regime, self-supervision does not complement natural language supervision.}
    \label{tab:large_imagenet}
    \vspace{-0.2cm}
\end{table}

At small-scale, we showed that self-supervision combined with natural language supervision marginally improves the quality of the learned visual representations, and that the choice of image-encoder pooling strategy can heavily influence results. We now investigate whether these conclusions still hold when training on a much larger, $1.4$B example dataset.

\subsection{Experimental Setup}
We follow the configuration described in Section \ref{sec:experimental_setup}, with a few key differences, noted below.

\textbf{Pretraining Dataset:} We combine a $2.2$B example web-crawled image-text pair dataset, termed the English-Web-Image-Text dataset (EWIT-$2.2$B), with LAION-$400$M \cite{schuhmann_laion-400m_2021}, CC$3$M \cite{sharma_gcc_2018}, CC$12$M \cite{changpinyo_conceptual_2021} and an internal high-quality image-text pair dataset containing approximately $134$M image text pairs which we term the High Quality Image Text Pairs Dataset (HQITP-134M). We globally deduplicate this---keyed by image bytes---reducing the number of image-text pairs significantly, to yield a final $1.4$B examples. Details provided in the supplementary material. In the rest of the paper we refer to this as the \emph{web-crawled} or large-scale dataset.

\textbf{Pretraining configuration:} We increase the learning rate warm-up to $1,000$ steps, and compute a local contrastive loss for the first $10,000$ steps, as during early training this improves the models rate of convergence. We train for a total of $480,000$ steps, which corresponds to $6$ full passes through the dataset. We do not increase the batch size, and use the same learning rate schedule as for \emph{CC}.

\textbf{Model variants:} We train M3AE as well as both GAP and MAX variants of CLIP and MAE-CLIP, having noted in the small-scale regime that these are the highest performing architectures and pre-training strategies. We do not train MAE at scale due to resource limitations, and because its performance was significantly lower than the models that utilize natural language supervision---a finding that is consistent with prior literature \cite{geng_m3ae_2022}.

\begin{figure*}
    \centering
    \begin{subfigure}{0.14\textwidth}
        \centering
        \caption{CLIP}
        \includegraphics[width=\textwidth]{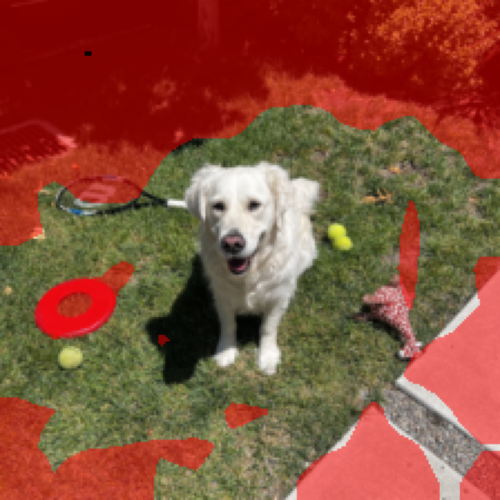}
    \end{subfigure}
    \begin{subfigure}{0.14\textwidth}
        \centering
        \caption{MAE-CLIP}
        \includegraphics[width=\textwidth]{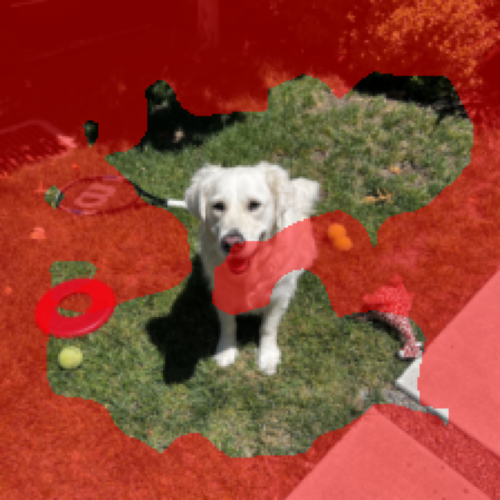}
    \end{subfigure}
    \begin{subfigure}{0.14\textwidth}
        \centering
        \caption{CLIP\textsubscript{GAP}}
        \includegraphics[width=\textwidth]{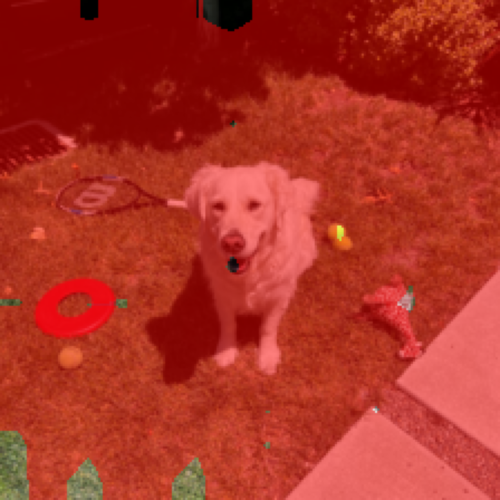}
    \end{subfigure}
    \begin{subfigure}{0.14\textwidth}
        \centering
        \caption{MAE-CLIP\textsubscript{GAP}}
        \includegraphics[width=\textwidth]{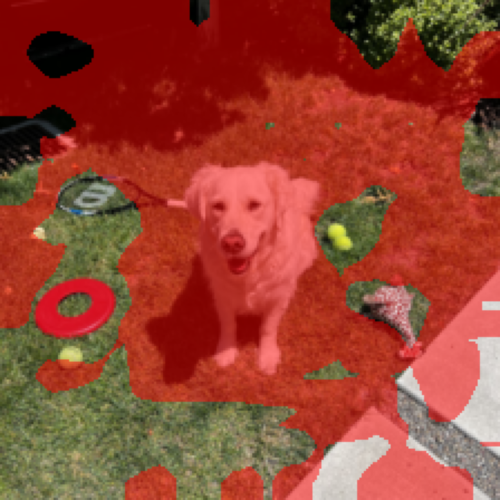}
    \end{subfigure}
    \begin{subfigure}{0.14\textwidth}
        \centering
        \caption{CLIP\textsubscript{MAX}}
        \includegraphics[width=\textwidth]{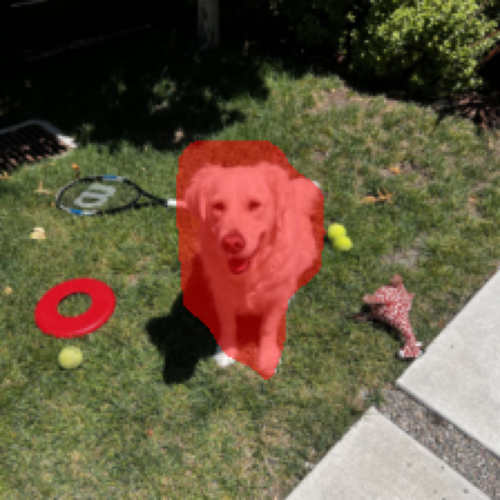}
    \end{subfigure}
    \begin{subfigure}{0.14\textwidth}
        \centering
        \caption{MAE-CLIP\textsubscript{MAX}}
        \includegraphics[width=\textwidth]{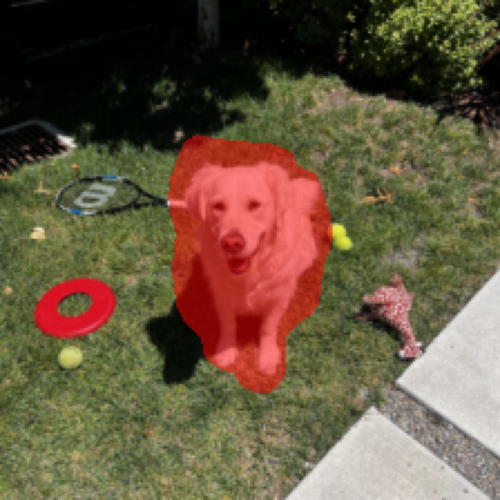}
    \end{subfigure}
    \begin{subfigure}{0.14\textwidth}
        \centering
        \includegraphics[width=\textwidth]{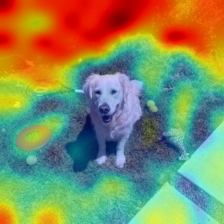}
    \end{subfigure}
    \begin{subfigure}{0.14\textwidth}
        \centering
        \includegraphics[width=\textwidth]{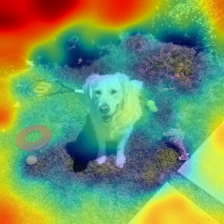}
    \end{subfigure}
    \begin{subfigure}{0.14\textwidth}
        \centering
        \includegraphics[width=\textwidth]{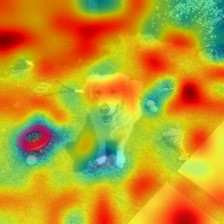}
    \end{subfigure}
    \begin{subfigure}{0.14\textwidth}
        \centering
        \includegraphics[width=\textwidth]{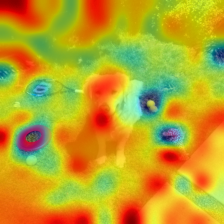}
    \end{subfigure}
    \begin{subfigure}{0.14\textwidth}
        \centering
        \includegraphics[width=\textwidth]{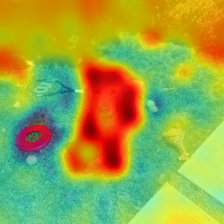}
    \end{subfigure}
    \begin{subfigure}{0.14\textwidth}
        \centering
        \includegraphics[width=\textwidth]{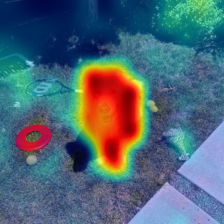}
    \end{subfigure}
    \caption{Zero-shot segmentation masks (top row) and GradCAM \cite{selvaraju_grad-cam_2020} visualizations (bottom row) for the prompt ``a photo of a dog'' for CLIP and MAE-CLIP. Self-supervision qualitatively improves the visual grounding of the model, however, the choice of pooling operator has the largest effect, with max pooling producing significantly better results \cite{ranasinghe2023perceptual}.}
    \label{fig:segmentation_visualization}
\end{figure*}

\subsection{Image Classification}
Table \ref{tab:large_imagenet} provides encoder performance for both zero-shot and linear-probing on ImageNet \cite{deng_imagenet_2009}. We note that at scale, MAE-CLIP typically shows worse performance than CLIP across both pooling strategies and tasks. We hypothesize that this may be due to the fact that masked-patch-prediction results in model capacity being lost to the task of predicting semantically irrelevant patches, meaning that we lose some of the benefits provided by the larger scale dataset (despite tuning the relative loss weights). If we are correct in this hypothesis then it is possible that a larger model capacity would help to alleviate this issue. We leave this to a future investigation.
Meanwhile, in Table \ref{tab:large_linear_probe} we once again explore the encoder performance through linear-probing for VTAB \cite{zhai_vtab_2020}. We note that GAP provides a consistent improvement over MAX across both CLIP and MAE-CLIP models, and that M3AE is also a strong performer---as was the case in the small-scale regime. M3AE provides particular benefit to the ``structured'' VTAB tasks, however underperforms (sometimes by a large margin) on the ``natural'' tasks. MAE-CLIP ends up in between the two---doing better than CLIP but worse than M3AE on the structured tasks, and worse than CLIP but better than M3AE on the natural tasks.

\subsection{VQA}
We follow the same finetuning strategy as before, with results in Table \ref{tab:large_vqa}. Notably, we do not see a consistent difference across tasks for CLIP vs MAE-CLIP, with the possible exception of CLEVR \cite{johnson_clevr_2017}, where it is clear that M3AE is by far the best performing approach---and its benefits improve the MAE-CLIP results on that task. GAP is perhaps a marginally stronger pooling strategy as measured against MAX, however the difference is smaller than it was in the small-scale regime.

\begin{table}
    \centering
    \begin{tabular}{lccc}
        \toprule
        \textbf{Model} & \textbf{CLEVR} & \textbf{VQAv2} & \textbf{GQA} \\
        \midrule
        M3AE         & \textbf{96.9} & 59.9 & 53.3 \\
        CLIP\textsubscript{GAP}     & 87.8 & \textbf{61.8} & 55.0 \\
        CLIP\textsubscript{MAX}     & 89.5 & 60.6 & 53.6 \\
        MAE-CLIP\textsubscript{GAP} & 92.8 & 61.9 & \textbf{55.3} \\ 
        MAE-CLIP\textsubscript{MAX} & 93.9 & 61.5 & 53.7 \\
        \bottomrule
    \end{tabular}
    \caption{VQA finetuning results on \emph{web-crawled} dataset (1.4B images). Similar to \Cref{tab:large_imagenet}, for large pretraining datasets, self-supervision is not improving the downstream VQA performance.}
    \label{tab:large_vqa}
    \vspace{-0.4cm}
\end{table}





\section{Analysis and Discussion}
\label{sec:analysis_discussion}

While our results present a pessimistic view on combining self-supervision with natural language supervision, we don't have a clear reason as to \emph{why} self-supervision fails at scale. Here we present two hypotheses that may be explored in future work.
\subsection{Visual Grounding}
\label{sec:grounding}
Visual grounding measures how well a particular representation or network can \emph{localize} objects within an image. It is thought that representations with better visual grounding will excel at more general purpose tasks. 
In section \ref{sec:exp}, we found that pooling operator had a larger effect on performance than the addition of MAE. We  qualitatively study the effect of both pooling and MAE on the visual grounding of the CLIP image and text encoders. We extract zero-shot segmentation masks and GradCAM \cite{selvaraju_grad-cam_2020} visualizations for the prompt ``a photo of a dog'', shown in \Cref{fig:segmentation_visualization}. Details are provided in the supplementary material. We observe that, with respect to visual grounding, the pooling operator plays a much more significant role than self-supervision \cite{ranasinghe2023perceptual}. In particular, CLIP\textsubscript{MAX} correctly segments the dog in the picture while CLIP focuses only in the background. However, both for CLIP and CLIP\textsubscript{MAX} we observe only an incremental improvement of the localization of the dog in the image when self-supervision is added. Unfortunately, even though the GAP variants of MAE-CLIP and CLIP achieve the best performance on VTAB, they do not also exhibit the most visual grounding. We hope that a more thorough future analysis can elucidate the relationship between self supervision and visual grounding.

\subsection{Dataset Diversity}
Alternatively, it may be that self-supervision and natural language supervision excel for entirely different parts of the dataset diversity-size spectrum.
Historically, the strongest self-supervised visual-encoder baselines (when compared to supervised methods) include methods such as: SimCLR \cite{simclr}, SimCLRV2 \cite{chen_simclrv2_2020} and masked auto encoders \cite{maes}, which all \emph{excel} on ImageNet---a somewhat diverse but a relatively ``clean" and object-centric dataset with 1000 disjoint classes. On less diverse datasets such as Cifar-10 \cite{cifar}, specialized self supervised methods such as ReMixMatch \cite{remixmatch} have shown excellent performance in the extremely low data regime. Outside of vision, MAEs have shown near state-of-the-art performance on datasets  such as Audioset \cite{gemmeke2017audio, maesaudio}, a clean 2M example, 632 class audio/video dataset. Recent works which perform controlled experiments scaling self supervised methods to massive datasets \cite{can} show that self-supervised methods achieve numbers roughly 10\% worse (as measured by ImageNet linear probe) as compared to natural language supervised methods at similar data scale.

On the other hand, natural language supervised models seem to only show competitive performance on massive and diverse datasets, CLIP \cite{clip} was trained on 400M examples, ALIGN \cite{align} was trained on 1.8B examples and BASIC \cite{basic} was trained on 6.6B examples. Recently \cite{noclip} performs controlled natural language supervision at a smaller scale, showing that natural language supervised models are 5-10\% worse (on ImageNet) than their supervised counterparts.  

While all of these past results only pose circumstantial and not rigorous experimental evidence, we believe it poses a fruitful line of future work on studying the scaling trends of self supervised methods and contrasting them with other forms of supervision.

\vspace{-0.1cm}
\section{Acknowledgement}
\label{sec:acknowledgement}
We are grateful to many colleagues for helpful advice and guidance throughout this project. We especially thank Jonathan Shlens, Brandon McKinzie, and Jason Ramuparam for feedback on paper drafts, Ruoming Pang for useful comments and suggestions early in the project, Tom Nickson for his training infrastructure support and Albin Madappally Jose for his assistance in building the training dataset. We also extend thanks to Dr. Nicola Strisciuglio for his clarifying input on earlier incarnations of this work, and Dr. Samuel Albanie for several engaging discussions and helpful pointers to related literature.

{\small
\bibliographystyle{ieee_fullname}
\bibliography{references}

\begin{thebibliography}{100}\itemsep=-1pt

\bibitem{alayrac_flamingo_2022}
Jean-Baptiste Alayrac, Jeff Donahue, Pauline Luc, Antoine Miech, Iain Barr,
  Yana Hasson, Karel Lenc, Arthur Mensch, Katie Millican, Malcolm Reynolds,
  Roman Ring, Eliza Rutherford, Serkan Cabi, Tengda Han, Zhitao Gong, Sina
  Samangooei, Marianne Monteiro, Jacob Menick, Sebastian Borgeaud, Andrew
  Brock, Aida Nematzadeh, Sahand Sharifzadeh, Mikolaj Binkowski, Ricardo
  Barreira, Oriol Vinyals, Andrew Zisserman, and Karen Simonyan.
\newblock Flamingo: a {Visual} {Language} {Model} for {Few}-{Shot} {Learning}.
\newblock Technical Report arXiv:2204.14198, arXiv, Apr. 2022.
\newblock arXiv:2204.14198 [cs] type: article.

\bibitem{arici_mlim_2021}
Tarik Arici, Mehmet~Saygin Seyfioglu, Tal Neiman, Yi Xu, Son Train, Trishul
  Chilimbi, Belinda Zeng, and Ismail Tutar.
\newblock {MLIM}: {Vision}-and-{Language} {Model} {Pre}-training with {Masked}
  {Language} and {Image} {Modeling}.
\newblock {\em arXiv:2109.12178 [cs]}, Sept. 2021.
\newblock arXiv: 2109.12178.

\bibitem{bao_beit_2021}
Hangbo Bao, Li Dong, and Furu Wei.
\newblock {BEiT}: {BERT} {Pre}-{Training} of {Image} {Transformers}.
\newblock {\em arXiv:2106.08254 [cs]}, June 2021.
\newblock arXiv: 2106.08254.

\bibitem{bao_vl-beit_2022}
Hangbo Bao, Wenhui Wang, Li Dong, and Furu Wei.
\newblock {VL}-{BEiT}: {Generative} {Vision}-{Language} {Pretraining}.
\newblock Technical Report arXiv:2206.01127, arXiv, June 2022.
\newblock arXiv:2206.01127 [cs] type: article.

\bibitem{remixmatch}
David Berthelot, Nicholas Carlini, Ekin~D. Cubuk, Alex Kurakin, Kihyuk Sohn,
  Han Zhang, and Colin Raffel.
\newblock Remixmatch: Semi-supervised learning with distribution alignment and
  augmentation anchoring.
\newblock {\em CoRR}, abs/1911.09785, 2019.

\bibitem{caron_dino_2021}
Mathilde Caron, Hugo Touvron, Ishan Misra, Hervé Jégou, Julien Mairal, Piotr
  Bojanowski, and Armand Joulin.
\newblock {DINO}: {Emerging} {Properties} in {Self}-{Supervised} {Vision}
  {Transformers}.
\newblock {\em arXiv:2104.14294 [cs]}, May 2021.
\newblock arXiv: 2104.14294.

\bibitem{changpinyo_conceptual_2021}
Soravit Changpinyo, Piyush Sharma, Nan Ding, and Radu Soricut.
\newblock Conceptual {12M}: {Pushing} {Web}-{Scale} {Image}-{Text}
  {Pre}-{Training} {To} {Recognize} {Long}-{Tail} {Visual} {Concepts}.
\newblock Technical Report arXiv:2102.08981, arXiv, Mar. 2021.
\newblock arXiv:2102.08981 [cs] type: article.

\bibitem{chen2021vseinfty}
Jiacheng Chen, Hexiang Hu, Hao Wu, Yuning Jiang, and Changhu Wang.
\newblock Learning the best pooling strategy for visual semantic embedding.
\newblock In {\em IEEE Conference on Computer Vision and Pattern Recognition
  (CVPR)}, 2021.

\bibitem{chen_igpt_2020}
Mark Chen, Alec Radford, Rewon Child, Jeffrey Wu, Heewoo Jun, David Luan, and
  Ilya Sutskever.
\newblock {iGPT}: {Generative} {Pretraining} {From} {Pixels}.
\newblock In {\em Proceedings of the 37th {International} {Conference} on
  {Machine} {Learning}}, pages 1691--1703. PMLR, Nov. 2020.
\newblock ISSN: 2640-3498.

\bibitem{chen_simclr_2020}
Ting Chen, Simon Kornblith, Mohammad Norouzi, and Geoffrey Hinton.
\newblock {SimCLR}: {A} {Simple} {Framework} for {Contrastive} {Learning} of
  {Visual} {Representations}.
\newblock In {\em Proceedings of the 37th {International} {Conference} on
  {Machine} {Learning}}, pages 1597--1607. PMLR, Nov. 2020.
\newblock ISSN: 2640-3498.

\bibitem{simclr}
Ting Chen, Simon Kornblith, Mohammad Norouzi, and Geoffrey~E. Hinton.
\newblock A simple framework for contrastive learning of visual
  representations.
\newblock {\em CoRR}, abs/2002.05709, 2020.

\bibitem{chen_simclrv2_2020}
Ting Chen, Simon Kornblith, Kevin Swersky, Mohammad Norouzi, and Geoffrey
  Hinton.
\newblock {SimCLRv2}: {Big} {Self}-{Supervised} {Models} are {Strong}
  {Semi}-{Supervised} {Learners}.
\newblock {\em arXiv:2006.10029 [cs, stat]}, Oct. 2020.
\newblock arXiv: 2006.10029.

\bibitem{chen_mocov2_2020}
Xinlei Chen, Haoqi Fan, Ross Girshick, and Kaiming He.
\newblock {MoCov2}: {Improved} {Baselines} with {Momentum} {Contrastive}
  {Learning}.
\newblock {\em arXiv:2003.04297 [cs]}, Mar. 2020.
\newblock arXiv: 2003.04297.

\bibitem{chen_simsiam_2020}
Xinlei Chen and Kaiming He.
\newblock {SimSiam}: {Exploring} {Simple} {Siamese} {Representation}
  {Learning}.
\newblock {\em arXiv:2011.10566 [cs]}, Nov. 2020.
\newblock arXiv: 2011.10566.

\bibitem{chen_mocov3_2021}
Xinlei Chen, Saining Xie, and Kaiming He.
\newblock {MoCov3}: {An} {Empirical} {Study} of {Training} {Self}-{Supervised}
  {Vision} {Transformers}.
\newblock {\em arXiv:2104.02057 [cs]}, Aug. 2021.
\newblock arXiv: 2104.02057.

\bibitem{chen_uniter_2020}
Yen-Chun Chen, Linjie Li, Licheng Yu, Ahmed El~Kholy, Faisal Ahmed, Zhe Gan, Yu
  Cheng, and Jingjing Liu.
\newblock {UNITER}: {UNiversal} {Image}-{TExt} {Representation} {Learning}.
\newblock In Andrea Vedaldi, Horst Bischof, Thomas Brox, and Jan-Michael Frahm,
  editors, {\em Computer {Vision} – {ECCV} 2020}, Lecture {Notes} in
  {Computer} {Science}, pages 104--120, Cham, Sept. 2020. Springer
  International Publishing.

\bibitem{clark_electra_2020}
Kevin Clark, Minh-Thang Luong, Quoc~V. Le, and Christopher~D. Manning.
\newblock {ELECTRA}: {Pre}-training {Text} {Encoders} as {Discriminators}
  {Rather} {Than} {Generators}.
\newblock Technical Report arXiv:2003.10555, arXiv, Mar. 2020.
\newblock arXiv:2003.10555 [cs] type: article.

\bibitem{imagenet}
Jia Deng, Wei Dong, Richard Socher, Li-Jia Li, Kai Li, and Li Fei-Fei.
\newblock Imagenet: A large-scale hierarchical image database.
\newblock In {\em 2009 IEEE conference on computer vision and pattern
  recognition}, pages 248--255. Ieee, 2009.

\bibitem{deng_imagenet_2009}
Jia Deng, Wei Dong, Richard Socher, Li-Jia Li, Kai Li, and Li Fei-Fei.
\newblock {ImageNet}: {A} large-scale hierarchical image database.
\newblock In {\em 2009 {IEEE} {Conference} on {Computer} {Vision} and {Pattern}
  {Recognition}}, pages 248--255, June 2009.
\newblock ISSN: 1063-6919.

\bibitem{desai_virtex_2021}
Karan Desai and Justin Johnson.
\newblock {VirTex}: {Learning} {Visual} {Representations} from {Textual}
  {Annotations}.
\newblock {\em arXiv:2006.06666 [cs]}, Sept. 2021.
\newblock arXiv: 2006.06666.

\bibitem{devlin_bert_2019}
Jacob Devlin, Ming-Wei Chang, Kenton Lee, and Kristina Toutanova.
\newblock {BERT}: {Pre}-training of {Deep} {Bidirectional} {Transformers} for
  {Language} {Understanding}.
\newblock {\em arXiv:1810.04805 [cs]}, May 2019.
\newblock arXiv: 1810.04805.

\bibitem{dosovitskiy_vit_2021}
Alexey Dosovitskiy, Lucas Beyer, Alexander Kolesnikov, Dirk Weissenborn,
  Xiaohua Zhai, Thomas Unterthiner, Mostafa Dehghani, Matthias Minderer, Georg
  Heigold, Sylvain Gelly, Jakob Uszkoreit, and Neil Houlsby.
\newblock {ViT}: {An} {Image} is {Worth} 16x16 {Words}: {Transformers} for
  {Image} {Recognition} at {Scale}.
\newblock {\em arXiv:2010.11929 [cs]}, June 2021.
\newblock arXiv: 2010.11929.

\bibitem{dou_meter_2021}
Zi-Yi Dou, Yichong Xu, Zhe Gan, Jianfeng Wang, Shuohang Wang, Lijuan Wang,
  Chenguang Zhu, Pengchuan Zhang, Lu Yuan, Nanyun Peng, Zicheng Liu, and
  Michael Zeng.
\newblock {METER}: {An} {Empirical} {Study} of {Training} {End}-to-{End}
  {Vision}-and-{Language} {Transformers}.
\newblock {\em arXiv:2111.02387 [cs]}, Nov. 2021.
\newblock arXiv: 2111.02387.

\bibitem{duan_multi-modal_2022}
Jiali Duan, Liqun Chen, Son Tran, Jinyu Yang, Yi Xu, Belinda Zeng, and Trishul
  Chilimbi.
\newblock Multi-modal {Alignment} using {Representation} {Codebook}.
\newblock {\em arXiv:2203.00048 [cs]}, Mar. 2022.
\newblock arXiv: 2203.00048.

\bibitem{el-nouby_splitmask_2021}
Alaaeldin El-Nouby, Gautier Izacard, Hugo Touvron, Ivan Laptev, Hervé Jegou,
  and Edouard Grave.
\newblock {SplitMask}: {Are} {Large}-scale {Datasets} {Necessary} for
  {Self}-{Supervised} {Pre}-training?
\newblock {\em arXiv:2112.10740 [cs]}, Dec. 2021.
\newblock arXiv: 2112.10740.

\bibitem{esser_vqgan_2021}
Patrick Esser, Robin Rombach, and Björn Ommer.
\newblock {VQGAN}: {Taming} {Transformers} for {High}-{Resolution} {Image}
  {Synthesis}.
\newblock {\em arXiv:2012.09841 [cs]}, June 2021.
\newblock arXiv: 2012.09841.

\bibitem{everingham_pascal_2010}
Mark Everingham, Luc Van~Gool, Christopher K.~I. Williams, John Winn, and
  Andrew Zisserman.
\newblock The {Pascal} {Visual} {Object} {Classes} ({VOC}) {Challenge}.
\newblock {\em International Journal of Computer Vision}, 88(2):303--338, June
  2010.

\bibitem{noclip}
Alex Fang, Gabriel Ilharco, Mitchell Wortsman, Yuhao Wan, Vaishaal Shankar,
  Achal Dave, and Ludwig Schmidt.
\newblock Data determines distributional robustness in contrastive language
  image pre-training (clip), 2022.

\bibitem{eva}
Yuxin Fang, Wen Wang, Binhui Xie, Quan Sun, Ledell Wu, Xinggang Wang, Tiejun
  Huang, Xinlong Wang, and Yue Cao.
\newblock Eva: Exploring the limits of masked visual representation learning at
  scale, 2022.

\bibitem{gemmeke2017audio}
Jort~F Gemmeke, Daniel~PW Ellis, Dylan Freedman, Aren Jansen, Wade Lawrence,
  R~Channing Moore, Manoj Plakal, and Marvin Ritter.
\newblock Audio set: An ontology and human-labeled dataset for audio events.
\newblock In {\em 2017 IEEE international conference on acoustics, speech and
  signal processing (ICASSP)}, pages 776--780. IEEE, 2017.

\bibitem{geng_m3ae_2022}
Xinyang Geng, Hao Liu, Lisa Lee, Dale Schuurmans, Sergey Levine, and Pieter
  Abbeel.
\newblock {M3AE}: {Multimodal} {Masked} {Autoencoders} {Learn} {Transferable}
  {Representations}.
\newblock Technical Report arXiv:2205.14204, arXiv, May 2022.
\newblock arXiv:2205.14204 [cs] type: article.

\bibitem{m3ae}
Xinyang Geng, Hao Liu, Lisa Lee, Dale Schuurmans, Sergey Levine, and Pieter
  Abbeel.
\newblock Multimodal masked autoencoders learn transferable representations,
  2022.

\bibitem{openwebtext_corpus}
Aaron Gokaslan and Vanya Cohen.
\newblock Openwebtext corpus.
\newblock \url{http://Skylion007.github.io/OpenWebTextCorpus}, 2019.

\bibitem{goyal_accurate_2018}
Priya Goyal, Piotr Dollár, Ross Girshick, Pieter Noordhuis, Lukasz Wesolowski,
  Aapo Kyrola, Andrew Tulloch, Yangqing Jia, and Kaiming He.
\newblock Accurate, {Large} {Minibatch} {SGD}: {Training} {ImageNet} in 1
  {Hour}.
\newblock {\em arXiv:1706.02677 [cs]}, Apr. 2018.
\newblock arXiv: 1706.02677.

\bibitem{goyal_vqa2_2017}
Yash Goyal, Tejas Khot, Douglas Summers-Stay, Dhruv Batra, and Devi Parikh.
\newblock {VQA2}: {Making} the v in {VQA} {Matter}: {Elevating} the {Role} of
  {Image} {Understanding} in {Visual} {Question} {Answering}.
\newblock pages 6904--6913, 2017.

\bibitem{grill_byol_2020}
Jean-Bastien Grill, Florian Strub, Florent Altché, Corentin Tallec, Pierre~H.
  Richemond, Elena Buchatskaya, Carl Doersch, Bernardo~Avila Pires,
  Zhaohan~Daniel Guo, Mohammad~Gheshlaghi Azar, Bilal Piot, Koray Kavukcuoglu,
  Rémi Munos, and Michal Valko.
\newblock {BYOL}: {Bootstrap} your own latent: {A} new approach to
  self-supervised {Learning}.
\newblock {\em arXiv:2006.07733 [cs, stat]}, Sept. 2020.
\newblock arXiv: 2006.07733.

\bibitem{maes}
Kaiming He, Xinlei Chen, Saining Xie, Yanghao Li, Piotr Doll{\'{a}}r, and
  Ross~B. Girshick.
\newblock Masked autoencoders are scalable vision learners.
\newblock {\em CoRR}, abs/2111.06377, 2021.

\bibitem{he_mae_2021}
Kaiming He, Xinlei Chen, Saining Xie, Yanghao Li, Piotr Dollár, and Ross
  Girshick.
\newblock {MAE}: {Masked} {Autoencoders} {Are} {Scalable} {Vision} {Learners}.
\newblock {\em arXiv:2111.06377 [cs]}, Nov. 2021.
\newblock arXiv: 2111.06377.

\bibitem{he_mocov1_2020}
Kaiming He, Haoqi Fan, Yuxin Wu, Saining Xie, and Ross Girshick.
\newblock {MoCov1}: {Momentum} {Contrast} for {Unsupervised} {Visual}
  {Representation} {Learning}.
\newblock pages 9729--9738, 2020.

\bibitem{Hu_2022_CVPR}
Shell~Xu Hu, Da Li, Jan St\"uhmer, Minyoung Kim, and Timothy~M. Hospedales.
\newblock Pushing the limits of simple pipelines for few-shot learning:
  External data and fine-tuning make a difference.
\newblock In {\em Proceedings of the IEEE/CVF Conference on Computer Vision and
  Pattern Recognition (CVPR)}, pages 9068--9077, June 2022.

\bibitem{maesaudio}
Po-Yao Huang, Hu Xu, Juncheng Li, Alexei Baevski, Michael Auli, Wojciech
  Galuba, Florian Metze, and Christoph Feichtenhofer.
\newblock Masked autoencoders that listen, 2022.

\bibitem{huang_seeing_2021}
Zhicheng Huang, Zhaoyang Zeng, Yupan Huang, Bei Liu, Dongmei Fu, and Jianlong
  Fu.
\newblock Seeing {Out} of the {Box}: {End}-to-{End} {Pre}-{Training} for
  {Vision}-{Language} {Representation} {Learning}.
\newblock pages 12976--12985, Apr. 2021.

\bibitem{hudson_gqa_2019}
Drew~A. Hudson and Christopher~D. Manning.
\newblock {GQA}: {A} {New} {Dataset} for {Real}-{World} {Visual} {Reasoning}
  and {Compositional} {Question} {Answering}.
\newblock pages 6700--6709, 2019.

\bibitem{align}
Chao Jia, Yinfei Yang, Ye Xia, Yi{-}Ting Chen, Zarana Parekh, Hieu Pham,
  Quoc~V. Le, Yun{-}Hsuan Sung, Zhen Li, and Tom Duerig.
\newblock Scaling up visual and vision-language representation learning with
  noisy text supervision.
\newblock {\em CoRR}, abs/2102.05918, 2021.

\bibitem{jia_align_2021}
Chao Jia, Yinfei Yang, Ye Xia, Yi-Ting Chen, Zarana Parekh, Hieu Pham, Quoc~V.
  Le, Yunhsuan Sung, Zhen Li, and Tom Duerig.
\newblock {ALIGN}: {Scaling} {Up} {Visual} and {Vision}-{Language}
  {Representation} {Learning} {With} {Noisy} {Text} {Supervision}.
\newblock {\em arXiv:2102.05918 [cs]}, June 2021.
\newblock arXiv: 2102.05918.

\bibitem{johnson_clevr_2017}
Justin Johnson, Bharath Hariharan, Laurens van~der Maaten, Li Fei-Fei,
  C.~Lawrence Zitnick, and Ross Girshick.
\newblock {CLEVR}: {A} {Diagnostic} {Dataset} for {Compositional} {Language}
  and {Elementary} {Visual} {Reasoning}.
\newblock In {\em 2017 {IEEE} {Conference} on {Computer} {Vision} and {Pattern}
  {Recognition} ({CVPR})}, pages 1988--1997, 2017.

\bibitem{kaggle_diabetic_retinopathy}
Kaggle and EyePacs.
\newblock Kaggle diabetic retinopathy detection.
\newblock \url{https://www.kaggle.com/c/diabetic-retinopathy-detection/data},
  July 2015.

\bibitem{kim_vilt_2021}
Wonjae Kim, Bokyung Son, and Ildoo Kim.
\newblock {ViLT}: {Vision}-and-{Language} {Transformer} {Without} {Convolution}
  or {Region} {Supervision}.
\newblock {\em arXiv:2102.03334 [cs, stat]}, June 2021.
\newblock arXiv: 2102.03334.

\bibitem{kingma_adam_2017}
Diederik~P. Kingma and Jimmy Ba.
\newblock Adam: {A} {Method} for {Stochastic} {Optimization}.
\newblock Technical Report arXiv:1412.6980, arXiv, Jan. 2017.
\newblock arXiv:1412.6980 [cs] type: article.

\bibitem{cifar}
Alex Krizhevsky, Geoffrey Hinton, et~al.
\newblock Learning multiple layers of features from tiny images.
\newblock 2009.

\bibitem{lewis_bart_2019}
Mike Lewis, Yinhan Liu, Naman Goyal, Marjan Ghazvininejad, Abdelrahman Mohamed,
  Omer Levy, Ves Stoyanov, and Luke Zettlemoyer.
\newblock {BART}: {Denoising} {Sequence}-to-{Sequence} {Pre}-training for
  {Natural} {Language} {Generation}, {Translation}, and {Comprehension}.
\newblock {\em arXiv:1910.13461 [cs, stat]}, Oct. 2019.
\newblock arXiv: 1910.13461.

\bibitem{li_semvlp_2021}
Chenliang Li, Ming Yan, Haiyang Xu, Fuli Luo, Wei Wang, Bin Bi, and Songfang
  Huang.
\newblock {SemVLP}: {Vision}-{Language} {Pre}-training by {Aligning}
  {Semantics} at {Multiple} {Levels}.
\newblock {\em arXiv:2103.07829 [cs]}, Mar. 2021.
\newblock arXiv: 2103.07829.

\bibitem{li_align_2021}
Junnan Li, Ramprasaath~R. Selvaraju, Akhilesh~Deepak Gotmare, Shafiq Joty,
  Caiming Xiong, and Steven Hoi.
\newblock Align before {Fuse}: {Vision} and {Language} {Representation}
  {Learning} with {Momentum} {Distillation}.
\newblock {\em arXiv:2107.07651 [cs]}, Oct. 2021.
\newblock arXiv: 2107.07651.

\bibitem{li_unimo-2_2022}
Wei Li, Can Gao, Guocheng Niu, Xinyan Xiao, Hao Liu, Jiachen Liu, Hua Wu, and
  Haifeng Wang.
\newblock {UNIMO}-2: {End}-to-{End} {Unified} {Vision}-{Language} {Grounded}
  {Learning}.
\newblock {\em arXiv:2203.09067 [cs]}, Mar. 2022.
\newblock arXiv: 2203.09067.

\bibitem{li_oscar_2020}
Xiujun Li, Xi Yin, Chunyuan Li, Pengchuan Zhang, Xiaowei Hu, Lei Zhang, Lijuan
  Wang, Houdong Hu, Li Dong, Furu Wei, Yejin Choi, and Jianfeng Gao.
\newblock Oscar: {Object}-{Semantics} {Aligned} {Pre}-training for
  {Vision}-{Language} {Tasks}.
\newblock In Andrea Vedaldi, Horst Bischof, Thomas Brox, and Jan-Michael Frahm,
  editors, {\em Computer {Vision} – {ECCV} 2020}, Lecture {Notes} in
  {Computer} {Science}, pages 121--137, Cham, July 2020. Springer International
  Publishing.

\bibitem{scalemask}
Yanghao Li, Haoqi Fan, Ronghang Hu, Christoph Feichtenhofer, and Kaiming He.
\newblock Scaling language-image pre-training via masking, 2022.

\bibitem{declip}
Yangguang Li, Feng Liang, Lichen Zhao, Yufeng Cui, Wanli Ouyang, Jing Shao,
  Fengwei Yu, and Junjie Yan.
\newblock Supervision exists everywhere: {A} data efficient contrastive
  language-image pre-training paradigm.
\newblock {\em CoRR}, abs/2110.05208, 2021.

\bibitem{liang_supmae_2022}
Feng Liang, Yangguang Li, and Diana Marculescu.
\newblock {SupMAE}: {Supervised} {Masked} {Autoencoders} {Are} {Efficient}
  {Vision} {Learners}.
\newblock Technical Report arXiv:2205.14540, arXiv, May 2022.
\newblock arXiv:2205.14540 [cs] type: article.

\bibitem{lin_2014_coco}
Tsung-Yi Lin, Michael Maire, Serge Belongie, James Hays, Pietro Perona, Deva
  Ramanan, Piotr Doll{\'a}r, and C~Lawrence Zitnick.
\newblock Microsoft coco: Common objects in context.
\newblock In {\em European conference on computer vision}, pages 740--755.
  Springer, 2014.

\bibitem{loshchilov_sgdr_2017}
Ilya Loshchilov and Frank Hutter.
\newblock {SGDR}: {Stochastic} {Gradient} {Descent} with {Warm} {Restarts}.
\newblock {\em arXiv:1608.03983 [cs, math]}, May 2017.
\newblock arXiv: 1608.03983.

\bibitem{loshchilov_adamw_2019}
Ilya Loshchilov and Frank Hutter.
\newblock {AdamW}: {Decoupled} {Weight} {Decay} {Regularization}.
\newblock {\em arXiv:1711.05101 [cs, math]}, Jan. 2019.
\newblock arXiv: 1711.05101.

\bibitem{lu_vilbert_2019}
Jiasen Lu, Dhruv Batra, Devi Parikh, and Stefan Lee.
\newblock {ViLBERT}: {Pretraining} {Task}-{Agnostic} {Visiolinguistic}
  {Representations} for {Vision}-and-{Language} {Tasks}.
\newblock {\em arXiv:1908.02265 [cs]}, Aug. 2019.
\newblock arXiv: 1908.02265.

\bibitem{can}
Shlok Mishra, Joshua Robinson, Huiwen Chang, David Jacobs, Aaron Sarna, Aaron
  Maschinot, and Dilip Krishnan.
\newblock A simple, efficient and scalable contrastive masked autoencoder for
  learning visual representations, 2022.

\bibitem{mu_slip_2021}
Norman Mu, Alexander Kirillov, David Wagner, and Saining Xie.
\newblock {SLIP}: {Self}-supervision meets {Language}-{Image} {Pre}-training.
\newblock {\em arXiv:2112.12750 [cs]}, Dec. 2021.
\newblock arXiv: 2112.12750.

\bibitem{slip}
Norman Mu, Alexander Kirillov, David~A. Wagner, and Saining Xie.
\newblock {SLIP:} self-supervision meets language-image pre-training.
\newblock {\em CoRR}, abs/2112.12750, 2021.

\bibitem{oord_representation_2019}
Aaron van~den Oord, Yazhe Li, and Oriol Vinyals.
\newblock Representation {Learning} with {Contrastive} {Predictive} {Coding}.
\newblock {\em arXiv:1807.03748 [cs, stat]}, Jan. 2019.
\newblock arXiv: 1807.03748.

\bibitem{basic}
Hieu Pham, Zihang Dai, Golnaz Ghiasi, Hanxiao Liu, Adams~Wei Yu, Minh{-}Thang
  Luong, Mingxing Tan, and Quoc~V. Le.
\newblock Combined scaling for zero-shot transfer learning.
\newblock {\em CoRR}, abs/2111.10050, 2021.

\bibitem{plummer_flickr30k_2015}
Bryan~A. Plummer, Liwei Wang, Chris~M. Cervantes, Juan~C. Caicedo, Julia
  Hockenmaier, and Svetlana Lazebnik.
\newblock Flickr30k {Entities}: {Collecting} {Region}-to-{Phrase}
  {Correspondences} for {Richer} {Image}-to-{Sentence} {Models}.
\newblock pages 2641--2649, 2015.

\bibitem{radford_clip_2021}
Alec Radford, Jong~Wook Kim, Chris Hallacy, Aditya Ramesh, Gabriel Goh,
  Sandhini Agarwal, Girish Sastry, Amanda Askell, Pamela Mishkin, Jack Clark,
  Gretchen Krueger, and Ilya Sutskever.
\newblock {CLIP}: {Learning} {Transferable} {Visual} {Models} {From} {Natural}
  {Language} {Supervision}.
\newblock {\em arXiv:2103.00020 [cs]}, Feb. 2021.
\newblock arXiv: 2103.00020.

\bibitem{clip}
Alec Radford, Jong~Wook Kim, Chris Hallacy, Aditya Ramesh, Gabriel Goh,
  Sandhini Agarwal, Girish Sastry, Amanda Askell, Pamela Mishkin, Jack Clark,
  Gretchen Krueger, and Ilya Sutskever.
\newblock Learning transferable visual models from natural language
  supervision.
\newblock {\em CoRR}, abs/2103.00020, 2021.

\bibitem{radford_gpt-2_2019}
Alec Radford, Jeffrey Wu, Rewon Child, David Luan, Dario Amodei, and Ilya
  Sutskever.
\newblock {GPT}-2: {Language} models are unsupervised multitask learners.
\newblock {\em OpenAI blog}, 1(8):9, 2019.

\bibitem{ranasinghe2023perceptual}
Kanchana Ranasinghe, Brandon McKinzie, Sachin Ravi, Yinfei Yang, Alexander~T
  Toshev, and Jonathon Shlens.
\newblock Perceptual grouping in vision-language models, 2023.

\bibitem{razavi_generating_2019}
Ali Razavi, Aaron van~den Oord, and Oriol Vinyals.
\newblock Generating {Diverse} {High}-{Fidelity} {Images} with {VQ}-{VAE}-2.
\newblock Technical Report arXiv:1906.00446, arXiv, June 2019.
\newblock arXiv:1906.00446 [cs, stat] type: article.

\bibitem{laion5b}
Christoph Schuhmann, Romain Beaumont, Richard Vencu, Cade Gordon, Ross
  Wightman, Mehdi Cherti, Theo Coombes, Aarush Katta, Clayton Mullis, Mitchell
  Wortsman, Patrick Schramowski, Srivatsa Kundurthy, Katherine Crowson, Ludwig
  Schmidt, Robert Kaczmarczyk, and Jenia Jitsev.
\newblock Laion-5b: An open large-scale dataset for training next generation
  image-text models, 2022.

\bibitem{schuhmann_laion-400m_2021}
Christoph Schuhmann, Richard Vencu, Romain Beaumont, Robert Kaczmarczyk,
  Clayton Mullis, Aarush Katta, Theo Coombes, Jenia Jitsev, and Aran
  Komatsuzaki.
\newblock {LAION}-{400M}: {Open} {Dataset} of {CLIP}-{Filtered} 400 {Million}
  {Image}-{Text} {Pairs}.
\newblock {\em arXiv:2111.02114 [cs]}, Nov. 2021.
\newblock arXiv: 2111.02114.

\bibitem{selvaraju_grad-cam_2020}
Ramprasaath~R. Selvaraju, Michael Cogswell, Abhishek Das, Ramakrishna Vedantam,
  Devi Parikh, and Dhruv Batra.
\newblock Grad-{CAM}: {Visual} {Explanations} from {Deep} {Networks} via
  {Gradient}-based {Localization}.
\newblock {\em International Journal of Computer Vision}, 128(2):336--359, Feb.
  2020.
\newblock arXiv:1610.02391 [cs].

\bibitem{sharma2018conceptual}
Piyush Sharma, Nan Ding, Sebastian Goodman, and Radu Soricut.
\newblock Conceptual captions: A cleaned, hypernymed, image alt-text dataset
  for automatic image captioning.
\newblock In {\em Proceedings of ACL}, 2018.

\bibitem{sharma_gcc_2018}
Piyush Sharma, Nan Ding, Sebastian Goodman, and Radu Soricut.
\newblock {GCC}: {Conceptual} {Captions}: {A} {Cleaned}, {Hypernymed}, {Image}
  {Alt}-text {Dataset} {For} {Automatic} {Image} {Captioning}.
\newblock In {\em Proceedings of the 56th {Annual} {Meeting} of the
  {Association} for {Computational} {Linguistics} ({Volume} 1: {Long}
  {Papers})}, pages 2556--2565, Melbourne, Australia, 2018. Association for
  Computational Linguistics.

\bibitem{flava}
Amanpreet Singh, Ronghang Hu, Vedanuj Goswami, Guillaume Couairon, Wojciech
  Galuba, Marcus Rohrbach, and Douwe Kiela.
\newblock {FLAVA:} {A} foundational language and vision alignment model.
\newblock {\em CoRR}, abs/2112.04482, 2021.

\bibitem{singh_flava_2022}
Amanpreet Singh, Ronghang Hu, Vedanuj Goswami, Guillaume Couairon, Wojciech
  Galuba, Marcus Rohrbach, and Douwe Kiela.
\newblock {FLAVA}: {A} {Foundational} {Language} {And} {Vision} {Alignment}
  {Model}.
\newblock {\em arXiv:2112.04482 [cs]}, Feb. 2022.
\newblock arXiv: 2112.04482.

\bibitem{instagram2}
Mannat Singh, Laura Gustafson, Aaron Adcock, Vinicius de Freitas~Reis, Bugra
  Gedik, Raj~Prateek Kosaraju, Dhruv Mahajan, Ross~B. Girshick, Piotr
  Doll{\'{a}}r, and Laurens van~der Maaten.
\newblock Revisiting weakly supervised pre-training of visual perception
  models.
\newblock {\em CoRR}, abs/2201.08371, 2022.

\bibitem{su_vl-bert_2020}
Weijie Su, Xizhou Zhu, Yue Cao, Bin Li, Lewei Lu, Furu Wei, and Jifeng Dai.
\newblock {VL}-{BERT}: {Pre}-training of {Generic} {Visual}-{Linguistic}
  {Representations}.
\newblock {\em arXiv:1908.08530 [cs]}, Feb. 2020.
\newblock arXiv: 1908.08530.

\bibitem{Sun_2017_ICCV}
Chen Sun, Abhinav Shrivastava, Saurabh Singh, and Abhinav Gupta.
\newblock Revisiting unreasonable effectiveness of data in deep learning era.
\newblock In {\em Proceedings of the IEEE International Conference on Computer
  Vision (ICCV)}, Oct 2017.

\bibitem{sun_amodal_2022}
Yihong Sun, Adam Kortylewski, and Alan Yuille.
\newblock Amodal {Segmentation} through {Out}-of-{Task} and
  {Out}-of-{Distribution} {Generalization} with a {Bayesian} {Model}.
\newblock Technical Report arXiv:2010.13175, arXiv, July 2022.
\newblock arXiv:2010.13175 [cs] type: article.

\bibitem{tan_lxmert_2019}
Hao Tan and Mohit Bansal.
\newblock {LXMERT}: {Learning} {Cross}-{Modality} {Encoder} {Representations}
  from {Transformers}.
\newblock {\em arXiv:1908.07490 [cs]}, Dec. 2019.
\newblock arXiv: 1908.07490.

\bibitem{van_der_maaten_visualizing_2008}
Laurens Van~der Maaten and Geoffrey Hinton.
\newblock Visualizing data using t-{SNE}.
\newblock {\em Journal of machine learning research}, 9(11), 2008.

\bibitem{vaswani_attention_2017}
Ashish Vaswani, Noam Shazeer, Niki Parmar, Jakob Uszkoreit, Llion Jones,
  Aidan~N. Gomez, {\textbackslash}Lukasz Kaiser, and Illia Polosukhin.
\newblock Attention is all you need.
\newblock In {\em Advances in neural information processing systems}, pages
  5998--6008, 2017.

\bibitem{vincent_extracting_2008}
Pascal Vincent, Hugo Larochelle, Yoshua Bengio, and Pierre-Antoine Manzagol.
\newblock Extracting and composing robust features with denoising autoencoders.
\newblock In {\em Proceedings of the 25th international conference on {Machine}
  learning - {ICML} '08}, pages 1096--1103, Helsinki, Finland, 2008. ACM Press.

\bibitem{beit}
Wenhui Wang, Hangbo Bao, Li Dong, Johan Bjorck, Zhiliang Peng, Qiang Liu, Kriti
  Aggarwal, Owais~Khan Mohammed, Saksham Singhal, Subhojit Som, and Furu Wei.
\newblock Image as a foreign language: Beit pretraining for all vision and
  vision-language tasks, 2022.

\bibitem{wang_unsupervised_2015}
Xiaolong Wang and Abhinav Gupta.
\newblock Unsupervised {Learning} of {Visual} {Representations} using {Videos}.
\newblock Technical Report arXiv:1505.00687, arXiv, Oct. 2015.
\newblock arXiv:1505.00687 [cs] type: article.

\bibitem{wang_simvlm_2021}
Zirui Wang, Jiahui Yu, Adams~Wei Yu, Zihang Dai, Yulia Tsvetkov, and Yuan Cao.
\newblock {SimVLM}: {Simple} {Visual} {Language} {Model} {Pretraining} with
  {Weak} {Supervision}.
\newblock {\em arXiv:2108.10904 [cs]}, Aug. 2021.
\newblock arXiv: 2108.10904.

\bibitem{wei_mvp_2022}
Longhui Wei, Lingxi Xie, Wengang Zhou, Houqiang Li, and Qi Tian.
\newblock {MVP}: {Multimodality}-guided {Visual} {Pre}-training.
\newblock {\em arXiv:2203.05175 [cs]}, Mar. 2022.
\newblock arXiv: 2203.05175.

\bibitem{xiao2010sun}
Jianxiong Xiao, James Hays, Krista~A Ehinger, Aude Oliva, and Antonio Torralba.
\newblock Sun database: Large-scale scene recognition from abbey to zoo.
\newblock In {\em 2010 IEEE computer society conference on computer vision and
  pattern recognition}, pages 3485--3492. IEEE, 2010.

\bibitem{xiao_early_2021}
Tete Xiao, Mannat Singh, Eric Mintun, Trevor Darrell, Piotr Dollár, and Ross
  Girshick.
\newblock Early {Convolutions} {Help} {Transformers} {See} {Better}.
\newblock Technical Report arXiv:2106.14881, arXiv, Oct. 2021.
\newblock arXiv:2106.14881 [cs] type: article.

\bibitem{xiong_pre-layernorm_2020}
Ruibin Xiong, Yunchang Yang, Di He, Kai Zheng, Shuxin Zheng, Chen Xing,
  Huishuai Zhang, Yanyan Lan, Liwei Wang, and Tieyan Liu.
\newblock Pre-layernorm: {On} {Layer} {Normalization} in the {Transformer}
  {Architecture}.
\newblock In {\em Proceedings of the 37th {International} {Conference} on
  {Machine} {Learning}}, pages 10524--10533. PMLR, Nov. 2020.
\newblock ISSN: 2640-3498.

\bibitem{focalnets}
Jianwei Yang, Chunyuan Li, Xiyang Dai, Lu Yuan, and Jianfeng Gao.
\newblock Focal modulation networks, 2022.

\bibitem{yao_filip_2021}
Lewei Yao, Runhui Huang, Lu Hou, Guansong Lu, Minzhe Niu, Hang Xu, Xiaodan
  Liang, Zhenguo Li, Xin Jiang, and Chunjing Xu.
\newblock {FILIP}: {Fine}-grained {Interactive} {Language}-{Image}
  {Pre}-{Training}.
\newblock {\em arXiv:2111.07783 [cs]}, Nov. 2021.
\newblock arXiv: 2111.07783.

\bibitem{yu_ernie-vil_2020}
Fei Yu, Jiji Tang, Weichong Yin, Yu Sun, Hao Tian, Hua Wu, and Haifeng Wang.
\newblock {ERNIE}-{ViL}: {Knowledge} {Enhanced} {Vision}-{Language}
  {Representations} {Through} {Scene} {Graph}.
\newblock June 2020.

\bibitem{coca}
Jiahui Yu, Zirui Wang, Vijay Vasudevan, Legg Yeung, Mojtaba Seyedhosseini, and
  Yonghui Wu.
\newblock Coca: Contrastive captioners are image-text foundation models, 2022.

\bibitem{yu_coca_2022}
Jiahui Yu, Zirui Wang, Vijay Vasudevan, Legg Yeung, Mojtaba Seyedhosseini, and
  Yonghui Wu.
\newblock {CoCa}: {Contrastive} {Captioners} are {Image}-{Text} {Foundation}
  {Models}.
\newblock {\em arXiv:2205.01917 [cs]}, May 2022.
\newblock arXiv: 2205.01917.

\bibitem{yuan_florence_2021}
Lu Yuan, Dongdong Chen, Yi-Ling Chen, Noel Codella, Xiyang Dai, Jianfeng Gao,
  Houdong Hu, Xuedong Huang, Boxin Li, Chunyuan Li, Ce Liu, Mengchen Liu,
  Zicheng Liu, Yumao Lu, Yu Shi, Lijuan Wang, Jianfeng Wang, Bin Xiao, Zhen
  Xiao, Jianwei Yang, Michael Zeng, Luowei Zhou, and Pengchuan Zhang.
\newblock Florence: {A} {New} {Foundation} {Model} for {Computer} {Vision}.
\newblock {\em arXiv:2111.11432 [cs]}, Nov. 2021.
\newblock arXiv: 2111.11432.

\bibitem{yun_patch-level_2022}
Sukmin Yun, Hankook Lee, Jaehyung Kim, and Jinwoo Shin.
\newblock Patch-level {Representation} {Learning} for {Self}-supervised
  {Vision} {Transformers}.
\newblock Technical Report arXiv:2206.07990, arXiv, June 2022.
\newblock arXiv:2206.07990 [cs] type: article.

\bibitem{zbontar_barlow_2021}
Jure Zbontar, Li Jing, Ishan Misra, Yann LeCun, and Stéphane Deny.
\newblock Barlow {Twins}: {Self}-{Supervised} {Learning} via {Redundancy}
  {Reduction}.
\newblock {\em arXiv:2103.03230 [cs, q-bio]}, June 2021.
\newblock arXiv: 2103.03230.

\bibitem{zeng_xvlm_2022}
Yan Zeng, Xinsong Zhang, and Hang Li.
\newblock X:{VLM}: {Multi}-{Grained} {Vision} {Language} {Pre}-{Training}:
  {Aligning} {Texts} with {Visual} {Concepts}.
\newblock {\em arXiv:2111.08276 [cs]}, Feb. 2022.
\newblock arXiv: 2111.08276.

\bibitem{zhai_2022_cvpr}
Xiaohua Zhai, Alexander Kolesnikov, Neil Houlsby, and Lucas Beyer.
\newblock Scaling vision transformers.
\newblock In {\em Proceedings of the IEEE/CVF Conference on Computer Vision and
  Pattern Recognition (CVPR)}, pages 12104--12113, June 2022.

\bibitem{vtab}
Xiaohua Zhai, Joan Puigcerver, Alexander Kolesnikov, Pierre Ruyssen, Carlos
  Riquelme, Mario Lucic, Josip Djolonga, Andr{\'{e}}~Susano Pinto, Maxim
  Neumann, Alexey Dosovitskiy, Lucas Beyer, Olivier Bachem, Michael Tschannen,
  Marcin Michalski, Olivier Bousquet, Sylvain Gelly, and Neil Houlsby.
\newblock The visual task adaptation benchmark.
\newblock {\em CoRR}, abs/1910.04867, 2019.

\bibitem{zhai_vtab_2020}
Xiaohua Zhai, Joan Puigcerver, Alexander Kolesnikov, Pierre Ruyssen, Carlos
  Riquelme, Mario Lucic, Josip Djolonga, Andre~Susano Pinto, Maxim Neumann,
  Alexey Dosovitskiy, Lucas Beyer, Olivier Bachem, Michael Tschannen, Marcin
  Michalski, Olivier Bousquet, Sylvain Gelly, and Neil Houlsby.
\newblock {VTAB}: {A} {Large}-scale {Study} of {Representation} {Learning} with
  the {Visual} {Task} {Adaptation} {Benchmark}.
\newblock {\em arXiv:1910.04867 [cs, stat]}, Feb. 2020.
\newblock arXiv: 1910.04867.

\bibitem{zhou_ade20k_2018}
Bolei Zhou, Hang Zhao, Xavier Puig, Tete Xiao, Sanja Fidler, Adela Barriuso,
  and Antonio Torralba.
\newblock {ADE20K}: {Semantic} {Understanding} of {Scenes} through the {ADE20K}
  {Dataset}.
\newblock {\em arXiv:1608.05442 [cs]}, Oct. 2018.
\newblock arXiv: 1608.05442.

\end{thebibliography}
}
\clearpage
\appendix
\newpage

\section{Supplementary Overview}
In what follows we first provide more exhaustive details on our training and evaluation datasets in Section \ref{sec:appendix:pre_training_datasets}, before then detailing our pre-training configuration in \ref{sec:appendix:training_configs}. Section \ref{sec:appendix:masking_influence} shows the effect of masking CLIP input in the low-data regime. Sections \ref{sec:appendix:retrieval}, \ref{sec:appendix:vtab_finetuning} and \ref{sec:tsne:analysis} show more results that were used to compare MAE-CLIP and CLIP. In Section \ref{sec:appendix:vqa_finetuning} we explain our VQA evaluation setup, and in Section \ref{sec:appendix:zs_seg} we cover zero-shot segmentation. Finally, Section \ref{sec:appendix:masking_strategy} gives background on our choice of ``similarity masking'' (as described in the main paper) over random masking, and Section \ref{sec:appendix:updated_m3ae_results} provides updated results for our large-scale M3AE \cite{geng_m3ae_2022} run, as at the time of submission we were unable to provide results for a fully trained baseline.
\newpage
\section{Datasets} \label{sec:appendix:pre_training_datasets}
In this section we describe our pre-training dataset, and also provide background as to why there are two missing tasks in our VTAB \cite{zhai_vtab_2020} evaluation.

\subsection{Pre-training Data}

We made use of two internal datasets as well as several public datasets to build our  ``large-scale'' pre-training dataset.

\textbf{High Quality Image Text Pairs dataset:}
The High Quality Image Text Pairs (HQITP-134M) dataset consists of approximately $134$M diverse and high quality images paired with descriptive captions and titles. Images range in spatial resolution from 320 to 2048 pixels on the short side. All images are JPEG format and most are RGB. Each example image is associated with a title, and a list of several captions. A small fraction ($\ll$ $1\%$) of the examples are missing both captions and title. We favor the associated captions, and find that these tokenize to an average length of $20.1$ tokens, although the shortest caption is only one token and the longest is over $1,000$. This dataset was licensed to our industrial research lab by a third party for commercial use.

\textbf{English Web Image Text dataset:}
The English-Web-Image-Text-2.2B (EWIT-2.2B) dataset consists of approximately $2.2$B images paired with one or more related pieces of text.  The data is the result of filtering English-language web-sourced data, using a combination of the filtering rules described in ALIGN \cite{jia_align_2021} and CLIP \cite{radford_clip_2021}.  Images range in spatial resolution from $200$ to $5000$ pixels on the short side, and $200$ to $8650$ on the long side, with a maximum aspect ratio of $3$ and a mean of $1.385$.  Each image has an average of $1.341$ pieces of text associated with it, although some have as many as $179$.  We find that the average associated text produces $15.5$ tokens when tokenized.

\textbf{Public datasets:}
As well as our internal datasets, we include Conceptual $12$M (CC12M) \cite{changpinyo_conceptual_2021}, CC3M \cite{sharma2018conceptual}, and LAION-$400$M \cite{schuhmann_laion-400m_2021}.  

\textbf{Overall pre-training dataset:}
Our overall training dataset is the result of combining HQITP-134M, CC12M, CC3M, and LAION-$400$M, before applying global image-byte-level de-duplication to drop image text pairs where either the image occurs more than once or the image occurs in one of our test sets.  This results in a final training dataset of just over $1.4$B image-text pairs.

\subsection{VTAB Evaluation Data}
In Sections 4 and 5 in the main paper we evaluate the quality of the learned visual representations by training a linear classifier on the predicted visual features of the VTAB datasets \cite{vtab}. However, we do not include the Diabetic Retinopathy \cite{kaggle_diabetic_retinopathy} dataset due to licensing concerns (the original dataset was provided solely for use in a Kaggle competition), and Sun397 \cite{xiao2010sun}, due to a missing image at the time of preparing the datasets for the VTAB benchmark. The issue with Sun397 has since then been resolved, and it could be included in a future iteration of this work.

\newpage
\section{Training configuration} \label{sec:appendix:training_configs}

\begin{table*}
\setlength{\tabcolsep}{2pt}
\small
\centering
\begin{tabular}{@{}llr@{}}
\toprule
\textbf{Component}             & \textbf{Parameter} & \multicolumn{1}{r}{\textbf{Value}}     \\ \midrule
Image Encoder                  & Depth              & 12                                     \\
\multirow{2}{*}{}              & Width              & 768                                   \\
                               & MLP Heads          & 12                                      \\ \midrule
Text Encoder                   & Depth              & 12                                     \\
\multirow{2}{*}{}              & Width              & 512                                    \\
                               & MLP Heads          & 8                                      \\ \midrule
Decoder                        & Depth              & 8                                      \\
                               & Width              & 512                                    \\
                               & MLP Heads          & 8                                      \\ \midrule
Model                          & Weight decay       & 0.1                                    \\
                               & Base LR            & 5e-04                               \\
                               & LR Schedule        & Cosine decay \cite{loshchilov_sgdr_2017}             \\
                               & LR Warmup steps        & 200             \\
                               & Local contrastive steps \cite{goyal_accurate_2018}       & 500                                  \\
                               & Batch size         & 256                                    \\
                               & Optimizer          & AdamW \cite{loshchilov_adamw_2019}              \\
                               & Optimizer momentum & $\beta$1 , $\beta$2 = 0.9, 0.98 \\
                               & Augmentation       & RandomResizedCrop  \\ \bottomrule
\end{tabular}
\caption{Parameters used for pre-training}
\label{tab:pretraining_parameters}
\end{table*}

\Cref{tab:pretraining_parameters} shows the used hyperparameter setup for pre-training of MAE-CLIP and all our baselines. Following \cite{radford_clip_2021}, we use an AdamW optimizer \cite{kingma_adam_2017, loshchilov_adamw_2019}, a linear learning rate warmup over $200$ steps before then decaying to $0$ with a cosine schedule \cite{loshchilov_sgdr_2017} over the remainder of training. Using warmup steps, (where only a local contrastive loss is used instead of a global contrastive loss), helps the model to converge faster at the beginning of training. To train MAE-CLIP, we simply sum the contrastive and generative losses for the local contrastive phase of training, but multiply the generative image loss by $0.05$ and the generative text loss by $0.1$ when computing the global contrastive loss. This allows us to accommodate for the dramatically reduced gradient norm of a global contrastive loss with a large batch size, and was arrived at through hyperparameter search. For the image encoder, initial empirical experimentation showed that using a trainable or fixed position encoding does not influence results and we therefore use a fixed 2D position encoding. For both image and text encoders, we use pre-layer-norm \cite{xiong_pre-layernorm_2020} and the initialization scheme from \cite{radford_gpt-2_2019}.

We use the same training configuration for the \emph{web-crawled} dataset, with a few changes due to the larger number of total steps. We use $10,000$ local contrastive loss steps, and $1,000$ warmup steps for the cosine learning rate scheduler.

\newpage
\section{Masking influence} \label{sec:appendix:masking_influence}
In MAE-CLIP, we never mask the input for the contrastive task. However, one can argue that in the low-data regime, masking is a heavy data augmentation that improves performance (similar to how e.g. DINO \cite{caron_dino_2021} has great performance in the low-data regime, likely also because of its heavy data augmentations). We therefore also run an ablation where the input to a contrastive-only (normal CLIP) model is masked, similar to how input is masked for the generative task in MAE-CLIP.

\begin{table}
    \centering
    \begin{tabular}{lcc}
    \toprule
        \textbf{Models} & \textbf{Zero-shot} & \textbf{Linear Probing} \\ 
        \midrule
        masked CLIP\textsubscript{MAP} & 23.0 \textit{(29.7)} & 48.0 \textit{(52.6)} \\
        masked CLIP\textsubscript{GAP} & 23.6 \textit{(29.3)} & 51.7 \textit{(59.8)} \\
        MAE-CLIP   & \textbf{33.8} & \textbf{58.9} \\ 
        \bottomrule
    \end{tabular}
    \caption{ImageNet classification with zero-shot transfer or linear probing after pretraining on the \emph{CC} dataset. CLIP is trained on masked input, showing unmasked performance between brackets.}
    \label{tab:cc_imagenet_masked}
\end{table}
\section{Retrieval results} \label{sec:appendix:retrieval}
\Cref{tab:zero-shot-retrieval} shows zero-shot retrieval accuracy for three different datasets. We compare COCO \cite{lin_2014_coco}, FLICKR \cite{plummer_flickr30k_2015} and COCOAmodal \cite{sun_amodal_2022}. We chose COCOAmodal as a third retrieval evaluation set, as adding the masked auto-encoder might have provided MAE-CLIP a benefit over CLIP on occluded objects or non-object-centric datasets. We show that even on non-object-centric datasets, CLIP outperforms MAE-CLIP at scale.

\begin{table}
\centering
\resizebox{0.45\textwidth}{!}{
\begin{tabular}{llllllllll}
\hline\multirow{2}{*}{\textbf{Model}} & \multicolumn{2}{l}{\textbf{COCO}}  & \multicolumn{2}{l}{\textbf{FLICKR}} & \multicolumn{2}{l}{\textbf{COCOA}}\\
& \textbf{I$\rightarrow$T} & \textbf{T$\rightarrow$I} & \textbf{I$\rightarrow$T} & \textbf{T$\rightarrow$I} & T1 & T5\\ \hline
CLIP\textsubscript{GAP}&51.9&36.6&78.8&62.3&24.2&46.9\\ 
CLIP\textsubscript{MAX}&\textbf{55.3}&\textbf{39.0}&80.5&\textbf{65.3}&22.7&\textbf{51.6}\\ 
MAE-CLIP\textsubscript{GAP}&53.0&37.0&77.3&62.0&20.7&39.5\\ 
MAE-CLIP\textsubscript{MAX}&54.4&37.7&\textbf{81.2}&64.2&\textbf{24.6}&41.4\\ 
\bottomrule
\end{tabular}
}
\caption{Zero-shot (retrieval) accuracy (\%). All trained on our \emph{web-crawled} dataset (1.4B images). I$\rightarrow$T: Image to Text, T$\rightarrow$I: Text to Image.}
\label{tab:zero-shot-retrieval}
\end{table}
\newpage
\section{VTAB Fine-tuning} \label{sec:appendix:vtab_finetuning}
\begin{table*}
\small
\centering
\setlength{\tabcolsep}{3.2pt}
\begin{tabular}{@{}l|ccccccccccccccccc|c@{}}
&
  \rot{{\color{vtab_natural}$\bullet$} Caltech101} &
  \rot{{\color{vtab_natural}$\bullet$} CIFAR-100} &
  \rot{{\color{vtab_natural}$\bullet$} DTD} &
  \rot{{\color{vtab_natural}$\bullet$} Flowers102} &
  \rot{{\color{vtab_natural}$\bullet$} Pets} &
  \multicolumn{1}{c|}{\rot{{\color{vtab_natural}$\bullet$} SVHN}} &
  \rot{{\color{vtab_specialized}$\bullet$} EuroSAT} &
  \rot{{\color{vtab_specialized}$\bullet$} Camelyon} &
  \multicolumn{1}{c|}{\rot{{\color{vtab_specialized}$\bullet$} Resisc45}} &
  \rot{{\color{vtab_structured}$\bullet$} Clevr/Closest} &
  \rot{{\color{vtab_structured}$\bullet$} Clevr/Count} &
  \rot{{\color{vtab_structured}$\bullet$} DMLab} &
  \rot{{\color{vtab_structured}$\bullet$} dSprites/Ori} &
  \rot{{\color{vtab_structured}$\bullet$} dSprites/Loc} &
  \rot{{\color{vtab_structured}$\bullet$} KITTI/Dist} &
  \rot{{\color{vtab_structured}$\bullet$} sNORB/Azim} &
  \multicolumn{1}{c|}{\rot{{\color{vtab_structured}$\bullet$} sNORB/Elev}} &
  \rot{Average} \\
  \midrule
    MAE-CLIP\textsubscript{GAP}
        & 92.6 & 84.2 & 74.4 & 95.1 & 85.7 & 80.2 & 97.0 & 88.0 & 92.9
        & 87.1 & 84.3 & 71.0 & 96.4 & 100 & 46.8 & 99.0 & 95.0 & 86.5 \\
    MAE-CLIP\textsubscript{MAX}
        & 95.9 & 86.3 & 80.3 & 97.5 & 90.4 & 97.3 & 98.9 & 88.8 & 95.6
        & 87.1 & 84.1 & 72.7 & 96.1 & 100 & 53.4 & 99.3 & 96.6 & 89.4 \\
    CLIP\textsubscript{GAP}
        & 96.0 & 87.7 & 81.8 & 98.1 & 90.7 & 97.4 & 98.8 & 85.7 & 95.9
        & 87.8 & 80.0 & 72.5 & 96.5 & 100 & 51.0 & 98.7 & 93.6 & 88.9  \\
    CLIP\textsubscript{MAX}
        & 95.5 & 86.3 & 81.1 & 98.0 & 90.6 & 97.6 & 98.9 & 89.6 & 96.2
        & 87.6 & 80.1 & 73.6 & 96.2 & 100 & 52.9 & 99.4 & 95.9 & 89.4  \\
    MAE-CLIP\textsubscript{GAP} \emph{CCxM}
        & 91.9 & 80.9 & 70.5 & 91.2 & 82.7 & 96.3 & 98.5 & 88.0 & 95.5
        & 79.6 & 67.7 & 66.0 & 96.3 & 100 & 49.6 & 99.7 & 85.9 & 84.7 \\
    MAE-CLIP\textsubscript{MAX} \emph{CCxM}
        & 90.8 & 81.6 & 69.0 & 90.3 & 81.0 & 96.8 & 98.4 & 89.5 & 95.3
        & 69.0 & 74.3 & 67.4 & 956 & 100 & 48.0 & 98.7 & 83.6 & 84.1 \\
    CLIP\textsubscript{GAP} \emph{CCxM}
        & 91.1 & 81.3 & 69.7 & 91.8 & 81.1 & 96.2 & 98.1 & 85.9 & 95.3
        & 75.1 & 65.1 & 64.6 & 96.4 & 100 & 48.6 & 99.6 & 78.0 & 83.4  \\
    CLIP\textsubscript{MAX} \emph{CCxM}
        & 89.6 & 80.6 & 68.4 & 90.3 & 79.8 & 96.5 & 98.4 & 87.1 & 95.3
        & 78.3 & 72.2 & 64.0 & 96.0 & 100 & 47.3 & 99.8 & 76.3 & 83.5  \\
    \bottomrule
\end{tabular}
\caption{Full-finetuning accuracy (\%) on classification tasks. Models are all trained on our \emph{web-crawled} dataset (1.4B images) or CCxM when specified.\\ {\color{vtab_natural}$\bullet$} VTAB/natural, {\color{vtab_specialized}$\bullet$} VTAB/specialized and {\color{vtab_structured}$\bullet$} VTAB/structured.}
\label{tab:full_finetuning}
\end{table*}

As MAE is often used in a full-finetuning setting, we also show that fully-finetuning follows the linear-probing results (see \Cref{tab:full_finetuning}). Adding masked autoencoding still does not outperform a contrastive-only baseline in a large-scale training. When training on CCxM only, we see MAE-CLIP outperforming CLIP on average.

\newpage
\section{T-SNE analysis} \label{sec:tsne:analysis}
\begin{figure*}
\vspace{-1.8cm}
  \centering
  \begin{subfigure}[b]{0.45\textwidth}
    \centering
    \includegraphics[width=\textwidth]{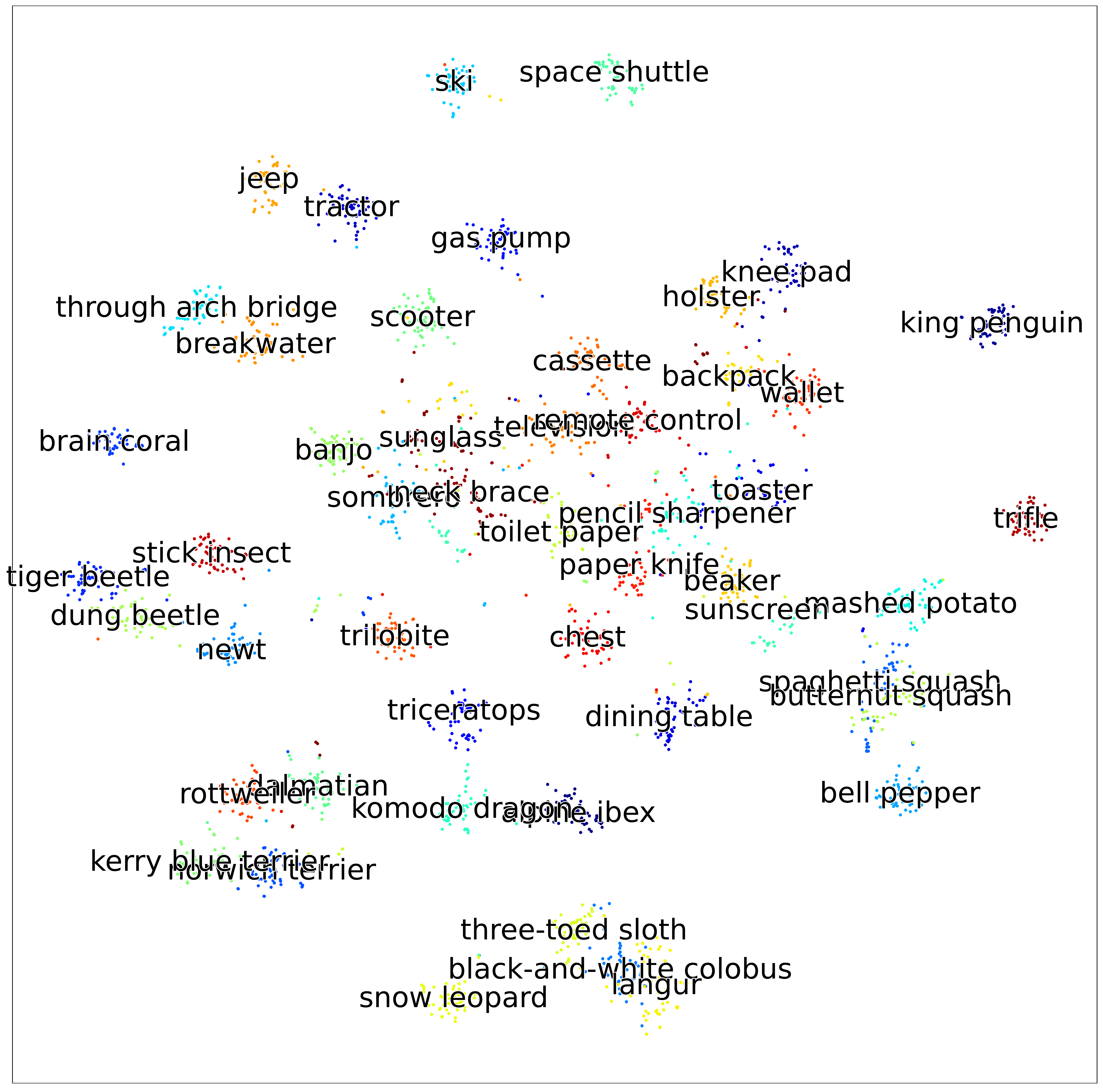}
    \caption{CLIP}
    \label{fig:tsne_clipgap}
  \end{subfigure}
  \begin{subfigure}[b]{0.45\textwidth}
    \centering
    \includegraphics[width=\textwidth]{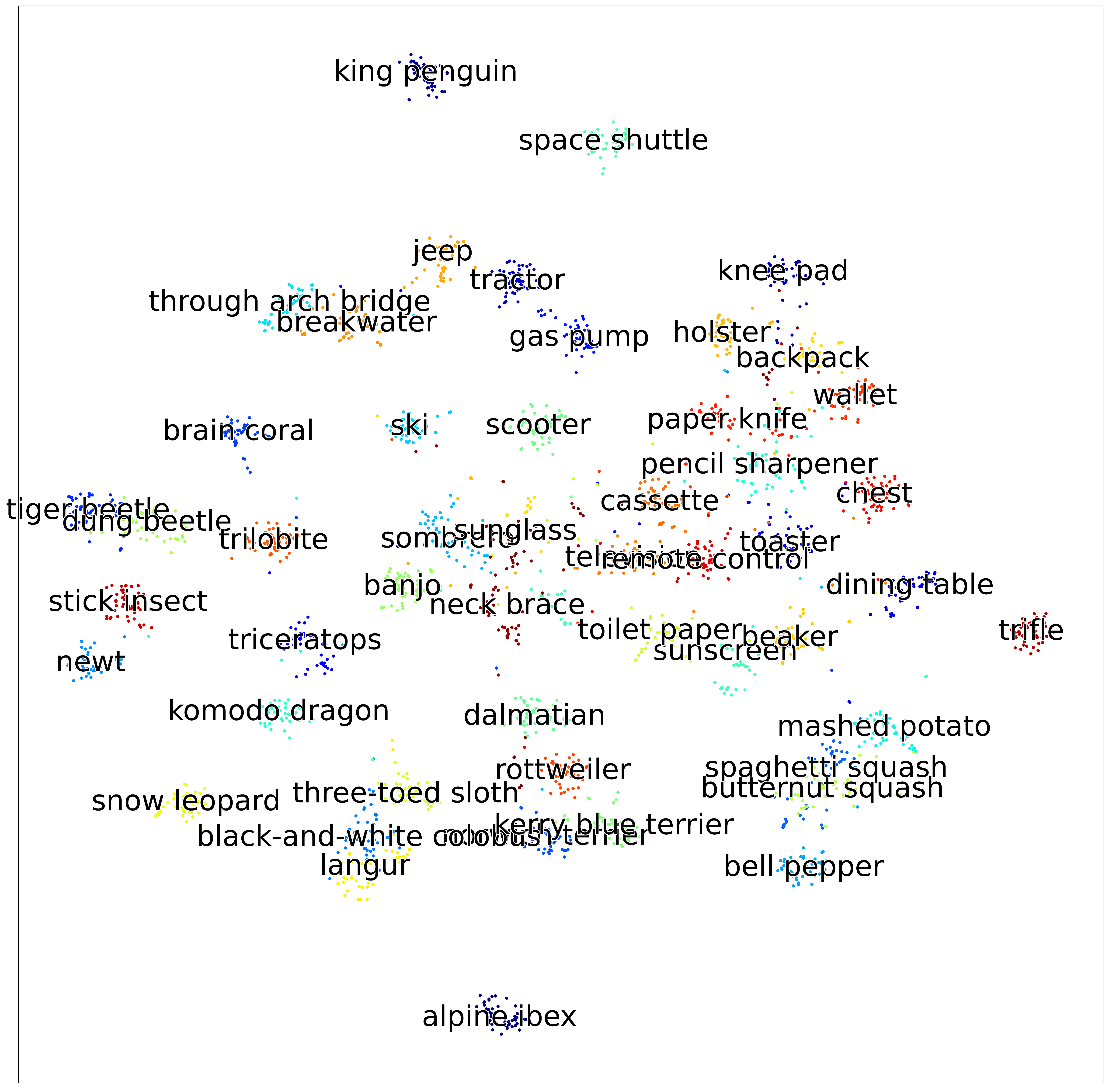}
    \caption{MAE-CLIP}
    \label{fig:tsne_maeclip}
  \end{subfigure}
  \begin{subfigure}[b]{0.45\textwidth}
    \centering
    \includegraphics[width=\textwidth]{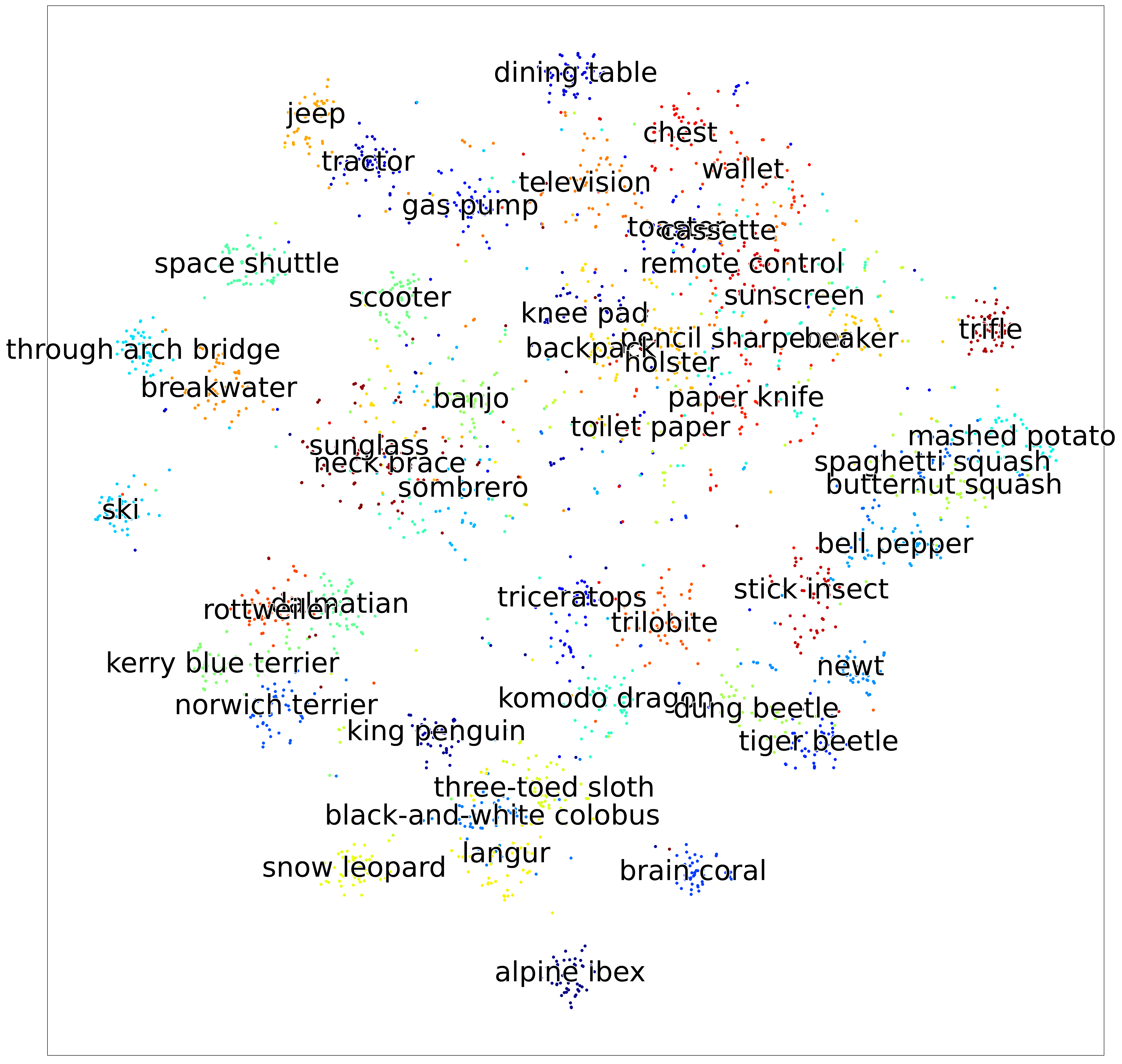}
    \caption{CLIP \emph{CCxM}}
    \label{fig:tsne_clipgapccxm}
  \end{subfigure}
  \begin{subfigure}[b]{0.45\textwidth}
    \centering
    \includegraphics[width=\textwidth]{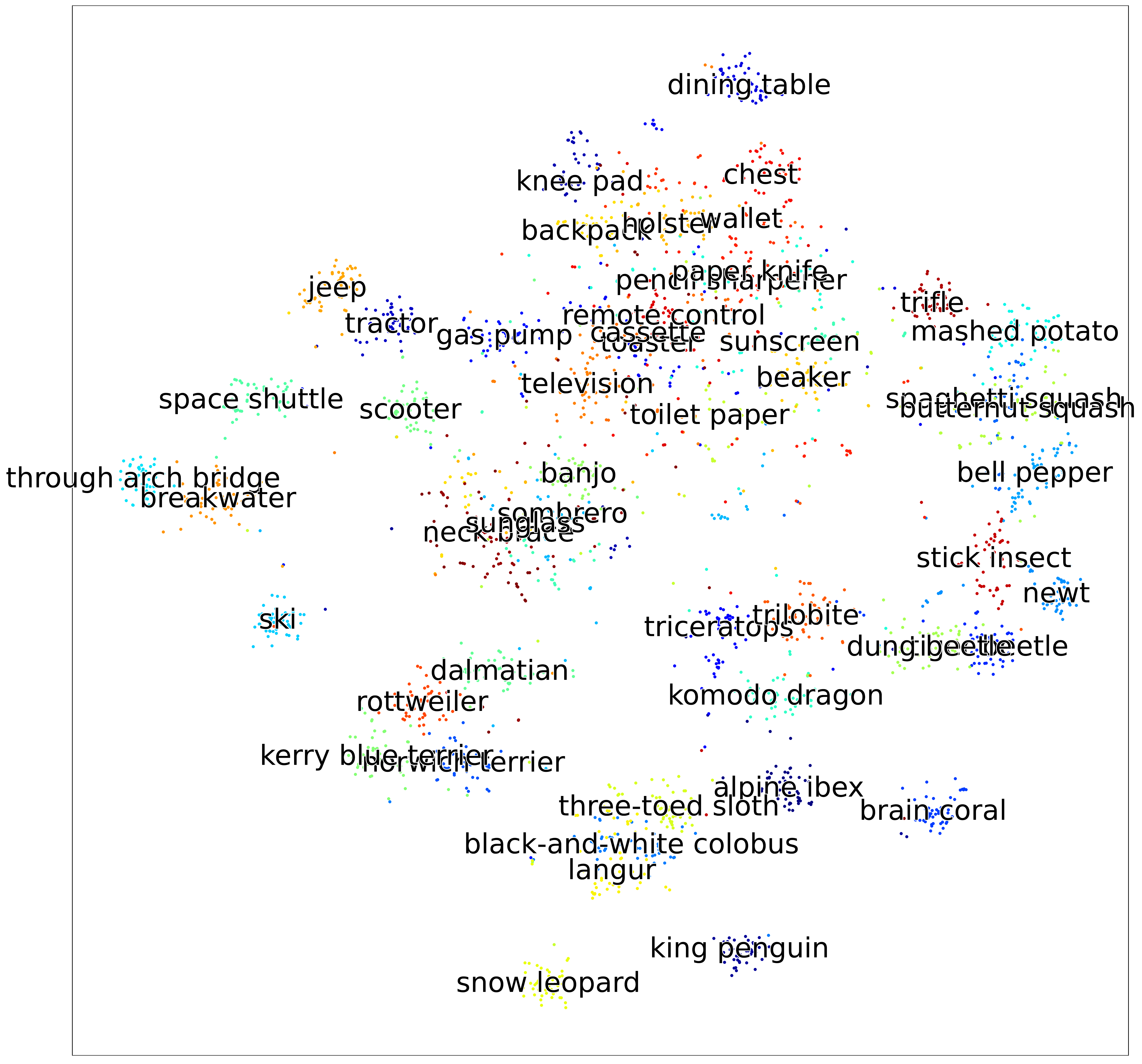}
    \caption{MAE-CLIP \emph{CCxM}}
    \label{fig:tsne_maeclipccxm}
  \end{subfigure}
  \begin{subfigure}[b]{0.45\textwidth}
    \centering
    \includegraphics[width=\textwidth]{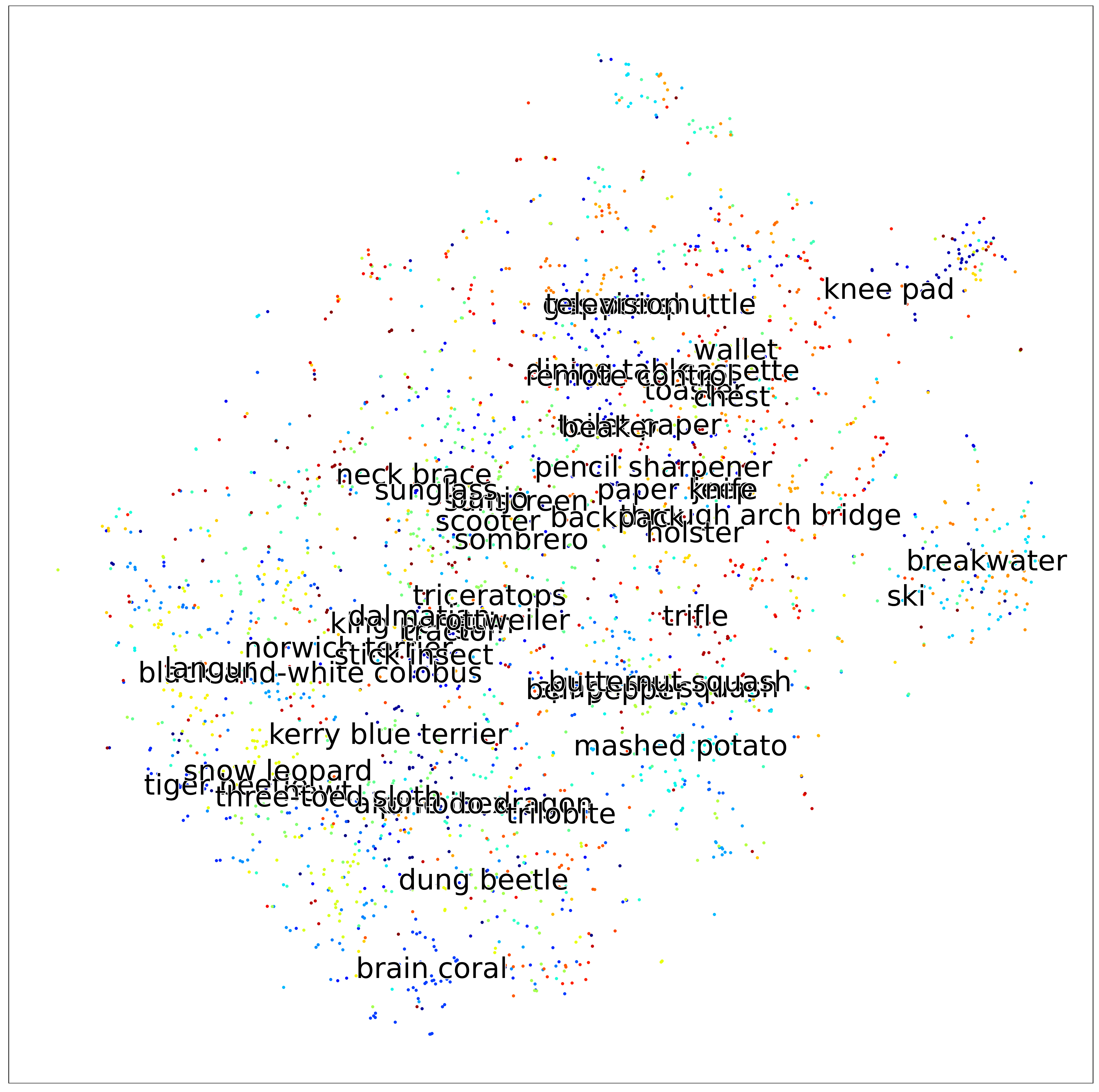}
    \caption{MAE \emph{CCxM}}
    \label{fig:tsne_maeccxm}
  \end{subfigure}
  \vspace{-0.3cm}
  \caption{T-sne visualizations for CLIP and MAE-CLIP, both large-scale training as CCXM.}
  \label{fig:tsne}
\end{figure*}

We also visually inspect the generated embeddings using t-SNE \cite{van_der_maaten_visualizing_2008}, see \Cref{fig:tsne}. There is no clear difference between the embeddings of CLIP and MAE-CLIP, at scale (see \Cref{fig:tsne_clipgap} and \Cref{fig:tsne_maeclip}). For CLIP, MAE-CLIP and MAE trained on CCMxM, we can see that the MAE embeddings (see \Cref{fig:tsne_maeccxm}) look more cluttered than the CLIP and MAE-CLIP embeddings.

\newpage
\section{VQA Finetuning} \label{sec:appendix:vqa_finetuning}
\begin{figure*}[h]
    \begin{center}
	\includegraphics[width=0.8\textwidth]{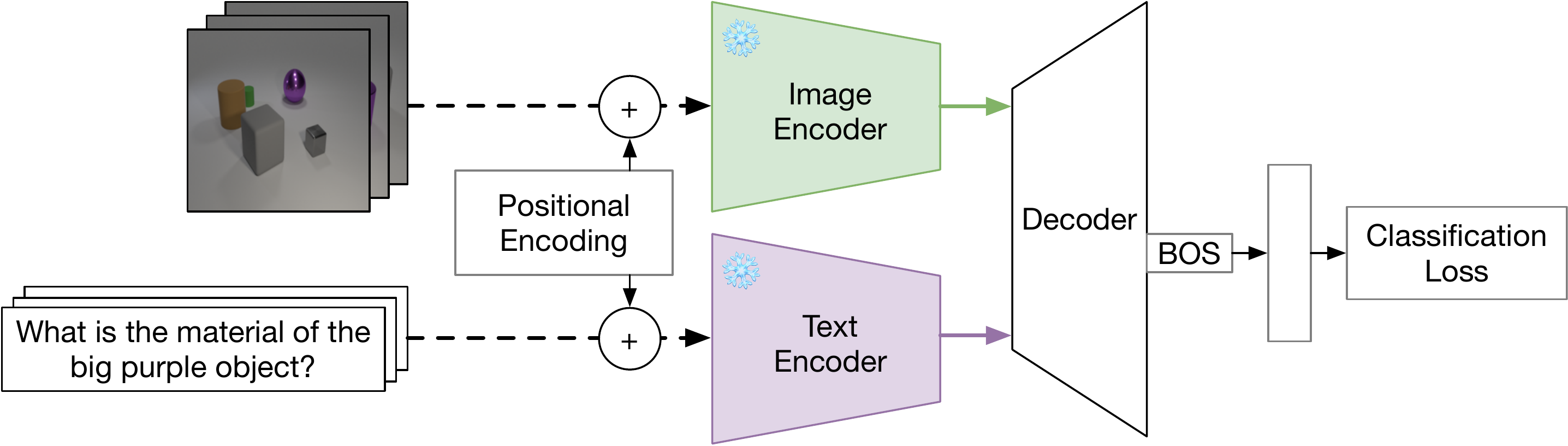}
    \end{center}
    \caption{VQA Fine-tuning: decoder classification. Encoders are frozen, the BOS token of the decoder output is used as the classification token.}
    \label{fig:vqa_decoderclassification}
\end{figure*}

We evaluate on three VQA benchmark datasets CLEVR \cite{johnson_clevr_2017}, VQAv2 \cite{goyal_vqa2_2017} and GQA \cite{hudson_gqa_2019}. As mentioned in the main paper, we finetune our models for VQA, by freezing the image and texts encoders and adding a new decoder. We treat the problem as a classification problem by calculating the set of possible answers and treating each as a separate class. Following our pre-training setup, we concatenate the image and text embeddings, add positional encoding and a modality specific token before using it as input in the decoder. The \verb+BOS+ token is used as an output token and linearly projected to the possible classes. \Cref{fig:vqa_decoderclassification} depicts our finetuning pipeline

\newpage
\section{Zero-shot Segmentation} \label{sec:appendix:zs_seg}
In Section 6 in the main paper, we present results on zero-shot semantic segmentation in order to evaluate the effects of self-supervision on visual grounding. Here, we describe our zero-shot semantic segmentation methodology and present a quantitative evaluation on three datasets. In particular, we use Pascal VOC \cite{everingham_pascal_2010}, ADE20K \cite{zhou_ade20k_2018} and COCO \cite{lin_2014_coco} with 20, 150 and 133 labels respectively. To compute segmentation masks, we first extract a feature per input pixel using bilinear interpolation from the per-patch features. Subsequently, we classify each pixel by computing the similarity of the feature to the embedding of the prompt for each class.

\begin{table}[h]
    \centering
    \begin{tabular}{lcccc}
        \toprule
        \textbf{Model} & \textbf{Pooling} & \textbf{COCO} & \textbf{ADE20K} & \textbf{Pascal VOC} \\
        \midrule
        CLIP & MAP & 8.4 & 4.6 & 19.1 \\
        MAE-CLIP & MAP & 9.1 & 5.7 & 20.6 \\
        \midrule
        CLIP & GAP & 7.6 & 4.1 & 19.4 \\
        MAE-CLIP & GAP & 8.5 & 5.5 & 20.4 \\
        \midrule
        CLIP & MAX & 16.5 & 9.4 & 36.6 \\
        MAE-CLIP & MAX & 17.8 & 11.1 & 36.9 \\
        \bottomrule
    \end{tabular}
    \caption{Zero-shot semantic segmentation results for CLIP and MAE-CLIP after training on the CC dataset ($11.3$M images). MAE-CLIP consistently improves upon CLIP for semantic segmentation regardless of the pooling strategy, as also seen qualitatively in Figure 2 in the main paper.}
    \label{tab:semantic_segmentation}
\end{table}

\Cref{tab:semantic_segmentation} evaluates the performance of CLIP and MAE-CLIP trained on the CC dataset with respect to mean intersection over union for all three datasets. We observe that self-supervision consistently improves the performance of CLIP for zero-shot semantic segmentation. However, as mentioned in Section 6 in the main paper, the choice of pooling operator has a much larger effect.

\subsection{Prompts}

In this section, we provide the prompts used for our zero-shot semantic segmentation experiments. Each prompt is made into a sentence by prepending ``a photo of a'' or ``an'' depending on whether the label starts with a vowel. For COCO \cite{lin_2014_coco} we simply use the 133 label names as they are provided by \url{https://github.com/cocodataset/panopticapi}. For Pascal VOC \cite{everingham_pascal_2010} we use
\begin{quote}
    \texttt{aeroplane, bicycle, bird, boat, bottle, bus, car, cat, chair, cow, table, dog, horse, motorbike, person, potted plant, sheep, sofa, train, television monitor}
\end{quote}
for classes $1$ to $20$ and
\begin{quote}
    \texttt{background, bag, bed, bench, book, building, cabinet, ceiling, cloth, computer, cup, door, fence, floor, flower, food, grass, ground, keyboard, light, mountain, mouse, curtain, platform, sign, plate, road, rock, shelves, sidewalk, sky, snow, bedclothes, track, tree, truck, wall, water, window, wood}
\end{quote}
for the 0-th class. Namely, if a pixel is classified as any of the latter categories it is considered to be a background pixel. Finally, for ADE20K \cite{zhou_ade20k_2018} we associate several prompots for each of the 150 categories. Subsequently, we compute the similarity of each pixel with each of the prompts and select the maximum similarity per category from all the associated prompts. The per-category prompts are as follows:
\begin{minipage}{2.1\linewidth}
    \vspace{-1.6cm}
    \begin{multicols}{5}
    \scriptsize
    \texttt{
    \setlength{\itemsep}{0em}
    \begin{enumerate}
        \itemsep0em
        \item wall, walls, brick wall, stone wall, interior wall
        \item building, buildings, edifice, edifices
        \item sky, clouds
        \item floor, flooring
        \item tree, trees
        \item ceiling
        \item road, route, street, roads, streets, routes
        \item bed, beds
        \item windowpane, window, windows
        \item grass, grass field
        \item cabinet, cabinets, wall mounted cabine
        \item sidewalk, pavement
        \item person, child, girl, boy, woman, man, people, children, girls, boys, women, men
        \item earth, ground
        \item door, double door, doors
        \item table, tables, tablecloth
        \item mountain, mount, mountains
        \item plant, flora, plant life, plants, bushes
        \item curtain, drape, drapery, mantle, pall
        \item chair, chairs
        \item car, automobile, cars
        \item water
        \item painting, picture, paintings, pictures, wallart, framed canvas
        \item sofa, couch, sofas, couches
        \item shelf, shelves
        \item house exterior
        \item sea, ocean
        \item mirror, mirrors
        \item rug, carpet, carpeting
        \item field
        \item armchair, armchairs
        \item seat, seats
        \item fence, fencing
        \item desk, desks
        \item rock, stone, rocks, stones
        \item wardrobe, closet, press, wardrobes, closets
        \item lamp, lamps
        \item bathtub, bathing tub, bath, tub
        \item railing, rail
        \item cushion, cushions
        \item pedestal
        \item box, boxes
        \item column, pillar
        \item signboard, sign, signboards, signs
        \item chest of drawers, chest, bureau, dresser
        \item counter
        \item sand
        \item sink
        \item skyscraper, skyscrapers
        \item fireplace, hearth, open fireplace
        \item refrigerator, icebox
        \item grandstand, covered stand
        \item path
        \item stairs, steps
        \item runway
        \item case, display case, showcase, vitrine
        \item pool table, billiard table, snooker table
        \item pillow, pillows
        \item screen door, shower door
        \item stairway, staircase
        \item river
        \item bridge, span
        \item bookcase
        \item window screen, door screen
        \item coffee table, cocktail table
        \item toilet, commode, crapper, potty
        \item flower, flowers
        \item book, books
        \item hill
        \item bench, benches
        \item countertop, counter top, worktop
        \item stove, kitchen stove, kitchen range, kitchen range, cooking stove
        \item palm tree, palm trees
        \item kitchen island
        \item computer, computing machine, computing device, data processor, electronic computer, information processing system
        \item swivel chair
        \item boat
        \item bar
        \item arcade machine, arcade machines
        \item hovel, hut, hutch, shack, shanty
        \item bus, autobus, double-decker, jitney, motorbus, motorcoach, omnibus, passenger vehicle
        \item towel
        \item light bulb, lightbulb, bulb, incandescent lamp, electric light, electric-light bulb
        \item truck, motortruck
        \item tower, towers
        \item chandelier, pendant, pendent
        \item awning, sunshade, sunblind
        \item streetlight, street lamp
        \item booth, cubicle, stall, kiosk
        \item television receiver, television, television set, tv, tv set
        \item airplane, aeroplane, airplanes, aeroplanes
        \item dirt track
        \item apparel, wearing apparel, dress, clothes
        \item pole
        \item land, soil
        \item bannister, banister, balustrade, balusters, handrail
        \item escalator, moving staircase, moving stairway
        \item ottoman, pouf, pouffe, puff, hassock
        \item bottle, bottles, water bottle
        \item buffet, sideboard
        \item poster, posting, placard, notice, bill, card
        \item stage
        \item van
        \item ship
        \item fountain
        \item conveyer belt, conveyor belt, conveyer, conveyor, transporter
        \item canopy
        \item washer, automatic washer, washing machine
        \item plaything, toy, toys
        \item swimming pool, swimming bath
        \item stool, stools
        \item barrel, cask, barrels, casks
        \item basket, handbasket
        \item waterfall, falls
        \item tent, collapsible shelter
        \item bag, bags, gift bag, paper bag
        \item minibike, motorbike
        \item cradle
        \item oven
        \item ball, balls
        \item food, solid food
        \item step, stair
        \item tank, storage tank
        \item trade name, brand name, brand, marque
        \item microwave, microwave oven
        \item plant pots, plant pot, flower pot, flowerpot, planter
        \item animal, animate being, dog, cat, horse, cow, sheep, zebra, girraffe, bird
        \item bicycle, bike
        \item lake
        \item dishwasher, dish washer, dishwashing machine
        \item projection screen
        \item blanket, cover
        \item sculpture, sculptures
        \item exhaust hood
        \item sconce, sconce lamp, sconce light
        \item vase, vases
        \item traffic light, traffic signal, traffic lights
        \item tray, trays
        \item ashcan, trash can, garbage can, wastebin, ash bin, ash-bin, ashbin, dustbin, trash barrel, trash bin
        \item ceiling fan, floor fan
        \item pier, wharf, wharfage, dock
        \item crt screen
        \item plate, plates
        \item monitor, monitoring device, monitors
        \item bulletin board, notice board
        \item shower
        \item radiator
        \item cup, cups, drinking glass, drinking glasses
        \item clock
        \item flag, flags
    \end{enumerate}
    }
    \end{multicols}
    \end{minipage}
    \clearpage
\newpage
\section{MAE-CLIP masking strategy} \label{sec:appendix:masking_strategy}
In Table \ref{tab:similarity_masking} we compare random masking to similarity masking for MAE-CLIP\textsubscript{MAX}, training on the CC dataset. Our experiments show that both strategies perform very similarly in all cases with random masking showing a small improvement for classification tasks while similarity masking an improvement on VQA and semantic segmentation, namely tasks that benefit from better visual grounding. All MAE-CLIP experiments in the main paper employ similarity masking. Further experiments are needed to properly evaluate the effect of the masking strategy across different scales and pooling methods.

\begin{table*}[h]
    \centering
    \begin{tabular}{lcccccc}
        \toprule
        \textbf{Model} & \textbf{Masking} & \textbf{VQA\textsubscript{Avg.}} & \textbf{SemSeg\textsubscript{Avg.}} & \textbf{VTAB\textsubscript{Avg}} & \textbf{IN1K\textsubscript{LP}} & \textbf{IN\textsubscript{ZS}} \\
        \midrule
        MAE-CLIP\textsubscript{MAX} & similarity & \textbf{68.53} & \textbf{21.95} & 69.14          & 63.16          & 35.2 \\
        MAE-CLIP\textsubscript{MAX} & random     & 68.44          & 21.27          & \textbf{69.92} & \textbf{63.46} & \textbf{35.4}          \\
        \bottomrule
    \end{tabular}
    \caption{Similarity vs random masking for MAE-CLIP\textsubscript{MAX} trained on the CC dataset. We show average VQA, zero-shot semantic segmentation (SemSeg), and VTAB results, as well as ImageNet1K (IN) Top-1 linear probe and zero-shot scores as measured on the validation set.}
    \label{tab:similarity_masking}
\end{table*}

\section{M3AE results} \label{sec:appendix:updated_m3ae_results}
This section provides an updated version of the ``large-scale'' tables of results from Section 5 in the main paper. They are presented in Tables \ref{tab:large_linear_probe}, \ref{tab:large_imagenet} and \ref{tab:large_vqa}. We include these because our M3AE \cite{geng_m3ae_2022} baseline had not fully converged at time of submission. All numbers except those associated with M3AE are identical to the ones presented in the main paper.

Firstly, we report the full linear probing evaluation on the VTAB tasks (\Cref{tab:large_linear_probe}). We observe that M3AE performs on par with MAE-CLIP\textsubscript{GAP}. Moreover, we note that M3AE performs consistently worse for all VTAB natural tasks while outperforming MAE-CLIP on VTAB strucuted tasks. This trend also continues with the rest of the results, where M3AE performs measurably worse on ImageNet linear-classification (\Cref{tab:large_imagenet}) while siginficantly better on CLEVR VQA (\Cref{tab:large_vqa}); the former being a very natural task and the latter being very structured.

\begin{table*}[h]
\caption{Linear probing accuracy (\%) on classification tasks. Models are all trained on our \emph{web-crawled} dataset (1.4B images). ({\color{vtab_natural}$\bullet$} VTAB/natural, {\color{vtab_specialized}$\bullet$} VTAB/specialized and {\color{vtab_structured}$\bullet$} VTAB/structured.) In the large scale pretraining regime, the difference between MAE-CLIP and CLIP is reduced to $<1\%$. This table provides the results for a fully trained M3AE compared to Table 5 in the main paper where M3AE had completed 50\% of the training steps.}
\label{tab:large_linear_probe}
\small
\centering
\setlength{\tabcolsep}{3.2pt}
\begin{tabular}{@{}l|ccccccccccccccccc|c@{}}
&
  \rot{{\color{vtab_natural}$\bullet$} Caltech101} &
  \rot{{\color{vtab_natural}$\bullet$} CIFAR-100} &
  \rot{{\color{vtab_natural}$\bullet$} DTD} &
  \rot{{\color{vtab_natural}$\bullet$} Flowers102} &
  \rot{{\color{vtab_natural}$\bullet$} Pets} &
  \multicolumn{1}{c|}{\rot{{\color{vtab_natural}$\bullet$} SVHN}} &
  \rot{{\color{vtab_specialized}$\bullet$} EuroSAT} &
  \rot{{\color{vtab_specialized}$\bullet$} Camelyon} &
  \multicolumn{1}{c|}{\rot{{\color{vtab_specialized}$\bullet$} Resisc45}} &
  \rot{{\color{vtab_structured}$\bullet$} Clevr/Closest} &
  \rot{{\color{vtab_structured}$\bullet$} Clevr/Count} &
  \rot{{\color{vtab_structured}$\bullet$} DMLab} &
  \rot{{\color{vtab_structured}$\bullet$} dSprites/Ori} &
  \rot{{\color{vtab_structured}$\bullet$} dSprites/Loc} &
  \rot{{\color{vtab_structured}$\bullet$} KITTI/Dist} &
  \rot{{\color{vtab_structured}$\bullet$} sNORB/Azim} &
  \multicolumn{1}{c|}{\rot{{\color{vtab_structured}$\bullet$} sNORB/Elev}} &
  \rot{Average} \\
   \midrule
    M3AE
        & 94.0 & 75.6 & 78.8 & 96.1 & 84.6 & 67.8 & 97.2 & 85.3 & 92.4
        & 65.4 & 75.9 & 52.0 & 57.4 & 79.2 & 51.1 & 40.6 & 68.3 & 74.2 \\
    CLIP
        & 94.9 & 78.4 & 80.0 & 97.3 & 86.9 & 59.0 & 94.1 & 82.3 & 92.7
        & 45.6 & 62.1 & 46.0 & 46.1 & 53.3 & 50.9 & 20.3 & 35.8 & 66.2  \\
    CLIP\textsubscript{MAX}
        & 96.1 & 81.0 & 80.9 & 97.3 & 89.9 & 65.7 & 96.0 & 83.2 & 94.1
        & 52.8 & 67.8 & 49.9 & 59.5 & 67.6 & 41.2 & 23.4 & 45.8 & 70.1 \\
    MAE-CLIP\textsubscript{MAX}
        & 95.8 & 79.2 & 81.5 & 96.8 & 88.2 & 62.1 & 95.8 & 81.8 & 93.0
        & 52.0 & 66.9 & 49.6 & 53.7 & 72.5 & 53.0 & 32.3 & 45.4 & 70.6 \\
    CLIP\textsubscript{GAP}
        & 95.8 & 80.5 & 81.6 & 97.6 & 88.7 & 66.0 & 97.0 & 84.4 & 93.3
        & 56.7 & 71.4 & 53.3 & 58.0 & 70.1 & 50.6 & 38.3 & 55.1 & 72.9  \\
    MAE-CLIP\textsubscript{GAP}
        & 95.4 & 79.3 & 82.2 & 97.4 & 88.6 & 72.8 & 96.6 & 84.5 & 93.5
        & 57.5 & 73.6 & 52.7 & 57.5 & 71.2 & 51.6 & 45.6 & 55.2 & 73.8 \\
    \bottomrule
\end{tabular}
\vspace{-0.2cm}
\end{table*}

\begin{table}
    \centering
    \begin{subtable}{0.4\textwidth}
        \centering
        \begin{tabular}{lcc}
        \toprule
            \textbf{Models} & \textbf{Zero-shot} & \textbf{Linear Probing} \\ 
            \midrule
            M3AE       &   -- & 71.5 \\
            CLIP\textsubscript{GAP}       & 61.8 & 75.9 \\
            CLIP\textsubscript{MAX}       & \textbf{63.7} & \textbf{77.5} \\
            MAE-CLIP\textsubscript{GAP} & 57.4 & 75.7 \\ 
            MAE-CLIP\textsubscript{MAX} & 60.9 & 76.6 \\
            \bottomrule
        \end{tabular}
        \caption{ImageNet classification}
        \label{tab:large_imagenet}
    \end{subtable}
    \hspace{1cm}
    \begin{subtable}{0.4\textwidth}
        \centering
        \begin{tabular}{lccc}
            \toprule
            \textbf{Model} & \textbf{CLEVR} & \textbf{VQAv2} & \textbf{GQA} \\
            \midrule
            M3AE         & \textbf{97.2} & 61.1 & 54.6 \\
            CLIP\textsubscript{GAP}     & 87.8 & 61.8 & 55.0 \\
            CLIP\textsubscript{MAX}     & 89.5 & 60.6 & 53.6 \\
            MAE-CLIP\textsubscript{GAP} & 92.8 & \textbf{61.9} & \textbf{55.3} \\ 
            MAE-CLIP\textsubscript{MAX} & 93.9 & 61.5 & 53.7 \\
            \bottomrule
        \end{tabular}
        \caption{VQA finetuning results}
        \label{tab:large_vqa}
    \end{subtable}
    \caption{ImageNet classification and VQA results after pretraining on \emph{web-crawled} dataset (1.4B images). In the large scale regime, self-supervision does not complement natural language supervision and all methods perform similarly on both tasks. This table provides updated results for M3AE (after it was fully trained) compared to Tables 6 and 7 in the main paper.}
\end{table}

\end{document}